\documentclass{article}

 \usepackage[eandd, final]{neurips_2026}
\usepackage[utf8]{inputenc} % allow utf-8 input
\usepackage[T1]{fontenc}    % use 8-bit T1 fonts
\usepackage{hyperref}       % hyperlinks
\usepackage{url}            % simple URL typesetting
\usepackage{booktabs}       % professional-quality tables
\usepackage{amsfonts}       % blackboard math symbols
\usepackage{nicefrac}       % compact symbols for 1/2, etc.
\usepackage{microtype}      % microtypography
\usepackage{xcolor}         % colors
\usepackage{graphicx}   % \includegraphics
\usepackage{wrapfig}    % wrapfigure
\usepackage{caption}    % \captionsetup
\usepackage{amsmath}
\usepackage{enumitem}
\usepackage{multirow}
\usepackage{adjustbox} 
\usepackage{pifont, makecell}
\usepackage{fontawesome5} % envelope

\definecolor{param_orange}{HTML}{E8925A}
\definecolor{state_green}{HTML}{6CA973}
\definecolor{infer_blue}{HTML}{6B9AC4}
\definecolor{gray_val}{HTML}{808080}
\definecolor{collapse_red}{HTML}{D9534F}
\definecolor{gain_green}{HTML}{45A049}

\title{What Drives Test-Time Adaptation for CLIP? A Controlled Empirical Study from a Update Perspective}

\author{%
  Jiazhen Huang\textsuperscript{1,*}, Xiao Chen\textsuperscript{1,*}, Zhiming Liu\textsuperscript{1,*}, Yaru Sun\textsuperscript{2}, Jingyan Jiang\textsuperscript{2,$\dagger$}, Zhi Wang\textsuperscript{1,$\dagger$} \\
        $^{1}$Shenzhen International Graduate School, Tsinghua University 
        $^{2}$Shenzhen Technology University \\
        \texttt{huangjiazhen1125@gmail.com},  \quad
	\texttt{chen-x25@mails.tsinghua.edu.cn}
}

\begin{document}

\maketitle

\renewcommand{\thefootnote}{\fnsymbol{footnote}}
\footnotetext[1]{Equal contribution.}
\footnotetext[2]{Co-corresponding authors: Jingyan Jiang (jiangjingyan@sztu.edu.cn) and Zhi Wang (wangzhi@sz.tsinghua.edu.cn)}

\begin{abstract}
    Vision-Language Models (VLMs) such as CLIP have become a standard backbone for open-vocabulary recognition, yet their zero-shot predictions remain vulnerable to distribution shifts encountered at deployment. 
    Test-Time Adaptation (TTA) has recently been extended to CLIP as a lightweight solution, leading to a rapidly growing body of TTA4CLIP methods. 
    However, empirical progress in this area has largely outpaced our understanding of what truly drives adaptation, where their gains originate, and under which shifts they remain reliable. 
    In this paper, we take a step back from the pursuit of state-of-the-art accuracy and conduct a systematic controlled study of TTA4CLIP. 
    We first organize existing methods into three unified paradigms according to \emph{what is updated} at test time. 
    We then introduce TTABC, an open-source TTA Benchmark for CLIP, which standardizes evaluation protocols and integrates more than 20 representative methods. 
    Our controlled empirical analysis focuses on three key areas. First, we determine the driving factors in parameter-based methods, revealing that adaptation gains are primarily driven by test-time evidence and reliable proxies rather than heavy optimization. Second, we explore evidence utilization beyond heavy parameter tuning, showing that competitive and efficient performance can be achieved through cross- or current-sample evidence and lightweight prototype updates. Finally, we demonstrate that there is no silver bullet for TTA: no single adaptation paradigm is universally optimal, and the preferred paradigm depends on the nature of shift. 
    We hope our benchmark and study provide a clearer understanding of the current TTA4CLIP landscape and establish a foundation for further research. 
    The project page is available at \url{https://walawalagoose.github.io/TTABC/}.
\end{abstract}

\section{Introduction}
Vision-language models (VLMs), exemplified by CLIP~\cite{CLIP}, have achieved remarkable success in open-vocabulary visual recognition~\cite{vlm_survey}. 
By aligning images and texts in a shared pre-trained representation space, CLIP enables zero-shot transfer across a wide range of downstream tasks~\cite{vlm_downstream_1, vlm_downstream_2}.
However, despite its strong transferability, CLIP often struggles with downstream distribution shifts, which are common in real-world deployment~\cite{shift_1, shift_2}. 
This has motivated the extension of Test-Time Adaptation (TTA)~\cite{TTA_survey, TTAB, TTA_survey_2}, originally developed for closed-set vision models~\cite{Tent, ttd}, to CLIP as a lightweight inference-time technique~\cite{tpt, tda}.
We refer to this setting as TTA4CLIP. 

Existing TTA4CLIP methods have reported strong performance, yet empirical progress has outpaced our collective understanding of what actually makes these methods effective. 
Within each methodological paradigm, new approaches typically improve the state-of-the-art on a limited set of out-of-distribution benchmarks. 
However, the more fundamental questions: \emph{what} truly drives adaptation, \emph{how} adaptation can be achieved without expensive computational costs, and \emph{whether} these methods are universally applicable across all shift scenarios, remain insufficiently examined.

In this work, we deliberately step back from the pursuit of the highest accuracy and instead investigate what TTA4CLIP is truly doing. 
Although existing TTA4CLIP methods differ substantially in implementation, we show that they can be systematically unified into three paradigms according to \emph{what is updated} at test time, including (i) parameter-based methods, (ii) state-based methods, and (iii) inference-based methods. 
The goal of this work is to provide a comprehensive understanding of the current landscape of TTA4CLIP methods while identifying critical open problems for future research. 
To this end, we introduce TTABC, an open-source Test-Time Adaptation Benchmark for CLIP, which provides rigorous evaluations, comprehensive analyses, and extensive baselines. 
Building on TTABC, we conduct a controlled empirical study of more than 20 representative methods spanning all three paradigms and covering a broad spectrum of distribution shifts.

\begin{wrapfigure}{r}{0.38\linewidth}
    \vspace{-18pt}  
    \centering
    \includegraphics[width=0.99\linewidth]{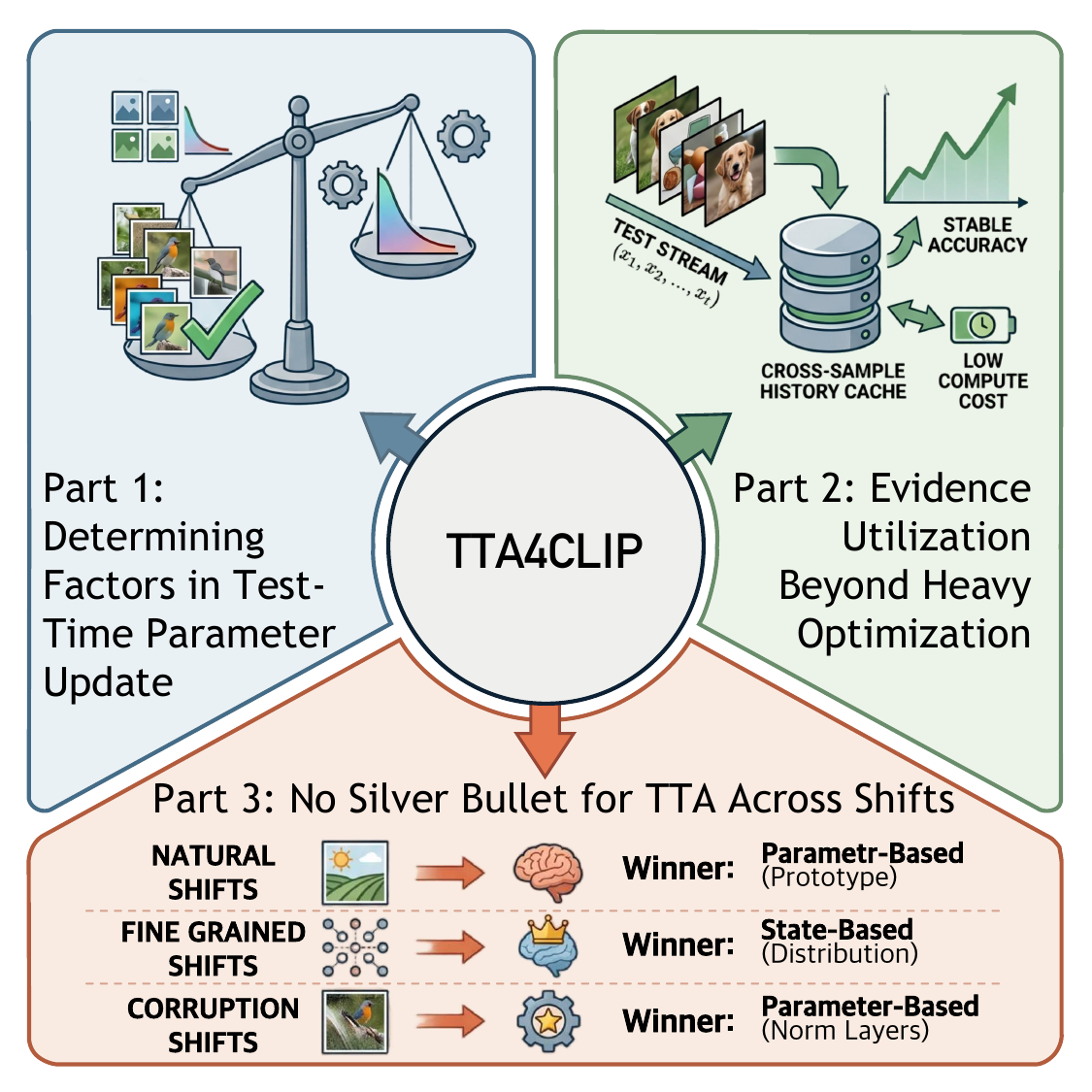}
    \vspace{-15pt}
    \captionsetup{width=0.99\linewidth}  % Set caption width to match image width
    \caption{\textbf{Overview of our controlled empirical study.} We investigate TTA4CLIP through three structured parts: (i) identifying the determining factors in parameter-based updates, (ii) exploring evidence utilization beyond heavy optimization, and (iii) evaluating different paradigms across diverse shifts.}
    \label{fig:intro}
    \vspace{-40pt}  
\end{wrapfigure}

Our study incorporates three key parts:
\begin{itemize} [leftmargin=*] 
    \item \textbf{Part 1: Determining factors in test-time parameter update.} Heavy parameter optimization yields limited and diminishing returns; instead, adaptation gains are primarily driven by the test-time evidence and reliable proxies.
    \item \textbf{Part 2: Evidence utilization beyond heavy optimization.} Competitive and efficient performance can be achieved through prediction refinement from cross/current-sample evidence or lightweight prototype residual updates, bypassing the need for heavy parameter optimization. 
    \item \textbf{Part 3: No silver bullet for TTA across shifts.} No single adaptation paradigm is universally optimal. The preferred paradigm largely depends on the nature of the shift.
\end{itemize}

Beyond these empirical findings, TTABC is designed as an extensible benchmark package that standardizes experimental protocols and facilitates the integration of new algorithmic implementations. We hope that this benchmark will enable more rigorous evaluation of TTA algorithms across a broader range of base models and distribution shifts, while also encouraging further research into the assumptions that determine when TTA is viable in challenging real-world scenarios.

\section{TTA4CLIP Taxonomy and Controlled Evaluation Setup}
\label{sec:settings}

Before presenting the empirical analyses that form the main contributions of this paper, we first fix notation and formalize the TTA setting in Sec.~\ref{sec:settings:prelim}, introducing our mechanistic taxonomy in Sec.~\ref{sec:settings:taxo}.

\subsection{Unified Formulation and Settings}
\label{sec:settings:prelim}

\paragraph{Classification with CLIP.}
A CLIP model contains an image encoder \(f_{\mathrm{I}}: \mathcal{X}\to\mathbb{R}^d\) and a text encoder \(f_{\mathrm{T}}: \mathcal{T}\to\mathbb{R}^d\), which map images and text into a shared embedding space. Given labels \(\mathcal{Y}=\{1,\ldots,K\}\) and a prompt template \(\tau(\cdot)\), each class prototype is
\(\mathbf{t}_y=f_{\mathrm{T}}(\tau(c_y))\). For a test image \(\mathbf{x}\), CLIP predicts
$
p_\theta(y \mid \mathbf{x}) =
\frac{\exp\!\left(\cos\langle f_{\mathrm{I}}(\mathbf{x}), \mathbf{t}_y\rangle/\gamma\right)}
{\sum_{y'=1}^{K}\exp\!\left(\cos\langle f_{\mathrm{I}}(\mathbf{x}), \mathbf{t}_{y'}\rangle/\gamma\right)},
$
where \(\gamma\) is the learned temperature and \(\theta=\{\theta_{\mathrm{I}},\theta_{\mathrm{T}}\}\). 
Zero-shot CLIP predicts independently for each sample without adaptation.

\paragraph{TTA4CLIP.}
Given the pre-trained model $\theta=\{\theta_{\mathrm{I}},\theta_{\mathrm{T}}\}$, at deployment, the model receives an unlabeled test stream
\(\mathcal{D}_{\mathrm{test}}=\{\mathbf{x}_t\}_{t=1}^{T}\), potentially shifted from the training distribution. 
Here, labels and source data are unavailable. 
The model is only allowed to adjust its prediction during inference given the current test sample or batch without external supervision. 
We write a general TTA step as:
\begin{equation}
(s_{t+1}, \phi_{t+1}) =
\mathcal{A}(s_t,\phi_t,\mathbf{x}_t;\theta),
\quad
\hat{y}_t =
g(s_t,\phi_t,\mathbf{x}_t;\theta),
\end{equation}
where $\mathcal{A}(\cdot)$ denotes the test-time update rule that modifies adaptable variables, and $g(\cdot)$ denotes the prediction rule producing the final output. 
\(\phi_t\) denotes the adaptable model parameters, such as prompts or norm layers; 
\(s_t\) denotes an external state, such as a memory bank or running statistics.

\paragraph{Our Proposed Evaluation Platform: TTABC.}
To make the empirical analyses interpretable, we propose TTABC, a benchmark serving as an evaluation platform that implements all methods within a shared codebase. 
We integrate \textbf{20+} TTA4CLIP methods spanning all three paradigms, with implementation details and hyperparameter configurations given in App.~\ref{app:detail}. 
We evaluate these methods across four diverse shift categories: 
(i) \emph{natural shifts}, including ImageNet-V2~\cite{inet-v2}, ImageNet-Sketch~\cite{inet-k}, ImageNet-A~\cite{inet-a}, and ImageNet-R~\cite{inet-r}; 
(ii) \emph{fine-grained categorizations}, encompassing StanfordCars~\cite{stanfordcars}, Food101~\cite{food101}, FGVC Aircraft~\cite{aircraft}, OxfordPets~\cite{pets}, Flowers102~\cite{flowers102}, SUN397~\cite{sun}, DTD~\cite{dtd}, EuroSAT~\cite{eurosat}, and UCF101~\cite{ucf101}; 
(iii) \emph{image corruptions}, evaluated on ImageNet-C~\cite{inet-c}; and 
(iv) \emph{label shifts}, which introduce temporal correlation into test stream and can be flexibly overlaid on any of the aforementioned types. 
Detailed descriptions of the datasets are provided in App.~\ref{app:base-data}.

\subsection{Mechanistic Taxonomy by Test-Time Update Target}
\label{sec:settings:taxo}
Classical TTA was primarily developed for closed-vocabulary vision models, with representative techniques including batch-normalization (BN) statistics re-estimation~\cite{bn_adapt} and entropy minimization (EM)~\cite{Tent, SAR, nctta, cliff, fan2026moetta}. 
These methods typically assume a monolithic classifier whose parameters can be safely adjusted on the target test stream. 
CLIP challenges this assumption due to its open-vocabulary and multimodal nature, which has prompted recent work on extending TTA to CLIP. 
Early attempts usually employ test-time tuning~\cite{tpt} inspired by few-shot transfer learning methods~\cite{coop, cocoop, maple, xiao2025visual,xiao2026layer}, adapting VLMs by optimizing learnable tokens or adapters. 
Later efforts increasingly turn to designing prediction rules for training-free adaptation, such as maintaining external online caches \cite{tda,dmn} or refining predictions from augmentations \cite{zero,mta}. 

Despite the apparent diversity of this literature, we argue that all existing TTA4CLIP methods can be systematically organized according to \emph{what is updated at test time}.
This perspective is still underexplored: a recent benchmark \cite{vlmtta} simply organizes them into episodic and online TTA, neglecting the essential differences among adaptation paradigms and protocols. 
For example, an episodic prompt-tuning method can also establish an online knowledge bank to accumulate historical memories \cite{dpe}, and the prompt can be adapted online \cite{histpt} as well.
Therefore, we establish a mechanistic taxonomy based on the \emph{test-time update target}, dividing TTA4CLIP methods into 3 categories, including: (i) \textbf{parameter-based}; (ii) \textbf{state-based} and (iii) \textbf{inference-based} methods. 

\begin{figure}[htbp]
    \centering
    \vspace{-10pt}
    \includegraphics[width=0.85\linewidth]{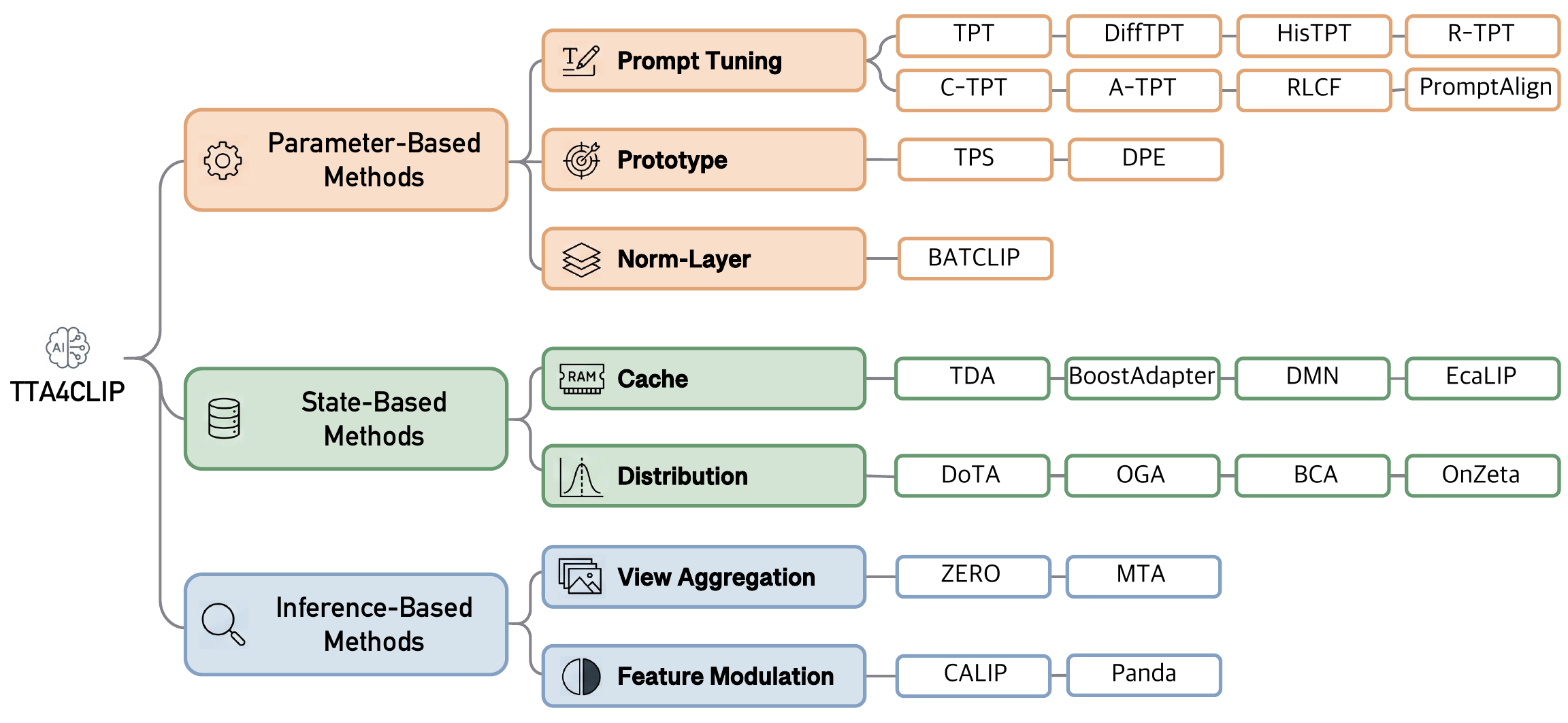}
    \vspace{-9pt}
    \caption{\textbf{A mechanistic taxonomy of TTA4CLIP.} Existing methods are categorized by their test-time update targets, including parameter-based, state-based, and inference-based methods.}
    \vspace{-15pt}
    \label{fig:taxonomy}
\end{figure}

\paragraph{Parameter-based.}
Parameter-based methods implement the most direct form of adaptation by optimizing an unsupervised objective derived from the test input to update test-time variables $\phi_t$:
\begin{equation}
\label{eq:para-update}
    \phi_{t+1} = \phi_t - \eta \nabla_{\phi} \mathcal{L}(\mathbf{x}_t; \phi_t, \theta),
    \qquad
    s_t \equiv s_0 .
\end{equation}
In this setting, $\phi_t$ denotes the adapted test-time parameters, such as prompt tokens, while $s_t$ denotes the external memory or cache state.
Here, adaptation is driven by gradient-based backpropagation optimization.
The most prominent branch is test-time prompt tuning. 
TPT~\cite{tpt} freezes both CLIP encoders, attaches learnable tokens to the text prompt, and minimizes marginal predicted entropy over filtered augmented views. 
Later approaches improve this template by enriching the view set, regularizing the optimization, or introducing auxiliary objectives. 
DiffTPT~\cite{difftpt} uses diffusion-generated views, while HisTPT~\cite{histpt} exploits historical prompts to improve stability during adaptation. 
C-TPT~\cite{ctpt} and A-TPT~\cite{atpt} constrain the geometry of class text embeddings for better calibration. 
R-TPT~\cite{rtpt} targets adversarial robustness, and PromptAlign~\cite{promptalign} and RLCF~\cite{rlcf} incorporate auxiliary alignment or reward signals. 
Another line of work modulates class prototypes rather than prompts. 
TPS~\cite{tps} learns per-class shifts for text prototypes, while DPE~\cite{dpe} co-evolves textual and visual prototypes over the test stream. 
More recent work such as BATCLIP~\cite{batclip} also adapts normalization or encoder parameters following classical TTA. 
% Although these methods differ in the updated variables, they share the same core mechanism: test-time evidence is converted into parameter updates through an unsupervised loss.
Most parameter-based methods require no persistent external state and thus are stateless. 
For exceptions like HisTPT and DPE, Eq. \eqref{eq:para-update} can be rewritten as:
$
    \phi_{t+1}
    =
    \phi_t - \eta \nabla_{\phi}\mathcal{L}(\mathbf{x}_t;\phi_t,\theta,s_t),
    % \quad
    s_{t+1} = \mathcal{U}(s_t, \mathbf{x}_t, \hat{\mathbf{y}}_t; \theta),
$
where $\mathcal{U}$ denotes the state-update rule.

\paragraph{State-based.}
State-based methods keep the adaptable parameters frozen, i.e., $\phi_t \equiv \phi_0$, and instead attach an external state $s_t$ that is updated through a specific read-write rule:
\begin{equation}
    s_{t+1} = \mathcal{U}(s_t, \mathbf{x}_t, \hat{\mathbf{y}}_t; \theta),
    \qquad
    \phi_t \equiv \phi_0 .
\end{equation}
In this paradigm, adaptation is driven by historical evidence accumulated from the test stream, rather than by gradient-based optimization. 
Cache-based methods implement this idea by storing and retrieving test features. 
TDA~\cite{tda} maintains class-wise high-confidence caches and combines zero-shot logits with nearest-neighbor scores, while using a negative cache to suppress confusing classes. 
BoostAdapter~\cite{boostadapter} enriches the cache via regional bootstrapping, DMN~\cite{dmn} separates short-term dynamic memory from long-term static memory for improved stability, and ECALP~\cite{ecalp} propagates labels over a graph of cached features. 
Other methods summarize the stream with compact distributional or relational statistics. 
DoTA~\cite{dota} and OGA~\cite{oga} maintain online Gaussian class densities, BCA~\cite{bca} performs Bayesian updating over class-conditional parameters and class frequencies, and OnZeta~\cite{onzeta} estimates the target class prior from streaming zero-shot predictions. 
% Despite different implementations, these methods share a common principle: they exploit historical test-time evidence through an external state while keeping the VLM parameters fixed.

\paragraph{Inference-based.}
While keeping the model frozen, inference-based methods make no modification of either parameters or external state.
Instead, they directly refine predictions within the forward pass:
\begin{equation}
    \phi_t \equiv \phi_0,
    \qquad
    s_t \equiv s_0,
    \qquad
    \hat{y}_t = g(\mathbf{x}_t; \theta).
\end{equation}
Here, adaptation is applied by using only currently available evidence such as augmented views, visual tokens or local patches. 
ZERO~\cite{zero} removes TPT's prompt optimization and relies on confidence-filtered voting over augmented views, while MTA~\cite{mta} uses mean-shift estimation to obtain a robust feature mode from the same view set. 
CALIP~\cite{calip} introduces parameter-free cross-modal attention between visual tokens and text prototypes, and Panda~\cite{panda} contrasts standard augmentations with negative ones to debias predictions.

Together, the three paradigms locate adaptation power in different places: parameter-based methods optimize test-time adaptable parameters, state-based methods accumulate historical evidence, and inference-based methods exploit current evidence. 
This taxonomy guides our later empirical analysis.

\section{Determining Factors in Test-Time Parameter Update}
\label{sec:tricks}
\begin{wrapfigure}{r}{0.5\linewidth}
    \vspace{-20pt} 
    \centering
    \includegraphics[width=0.99\linewidth]{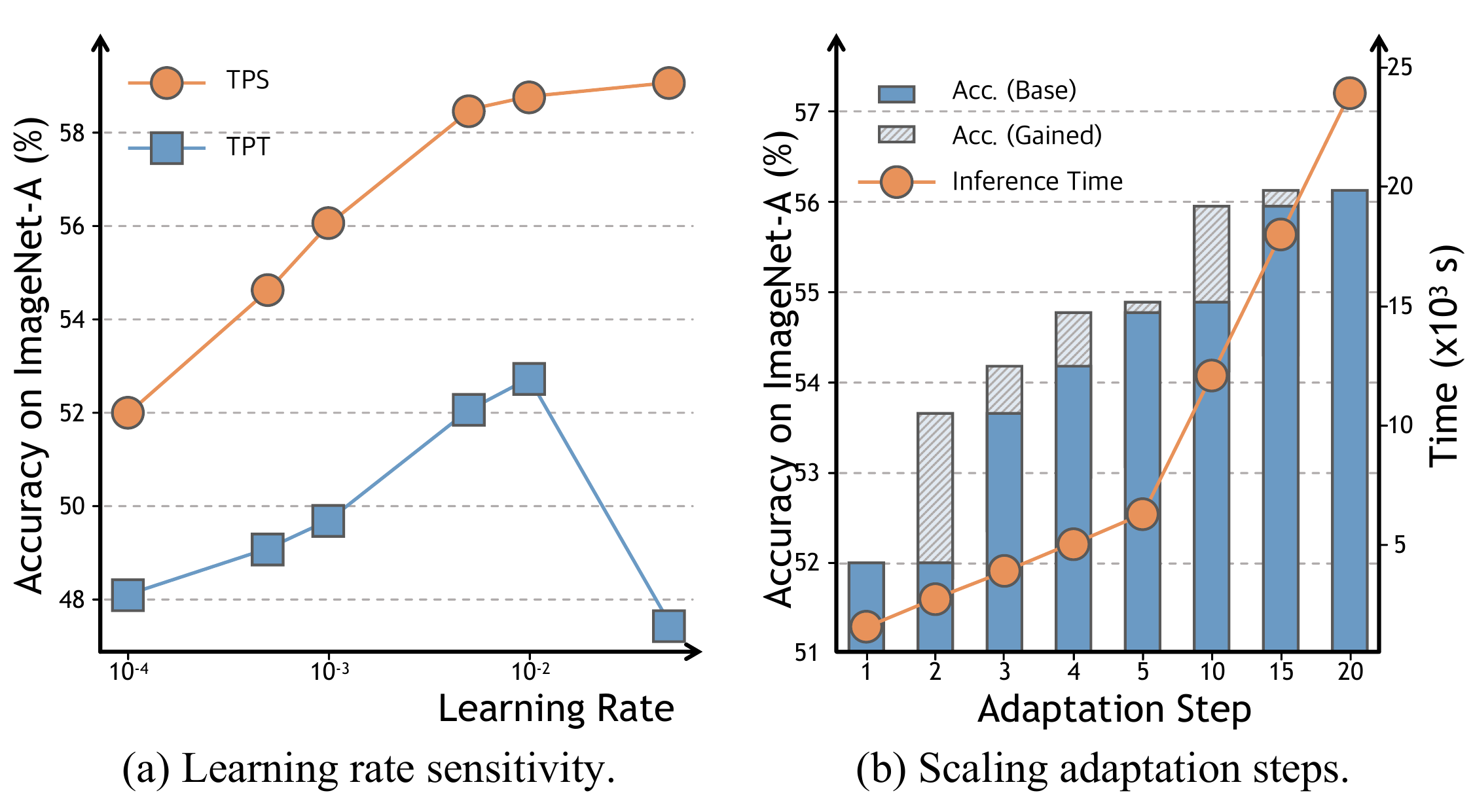}
    \vspace{-15pt}
    \captionsetup{width=0.99\linewidth}  % Set caption width to match image width
    % \caption{\textbf{Diminishing returns of heavy optimization.} Adjusting the learning rate exhibits high volatility and early saturation, while increasing per-sample adaptation steps incurs linear compute costs with rapidly marginal accuracy gains.}
    \caption{\textbf{Effect of Update Magnitude and Depth.} (a) Adjusting the learning rate causes accuracy fluctuations and leads to potentially suboptimal performance; (b) Increasing the adaptation steps yields limited and diminishing returns.}
    \label{fig:lr_step}
    \vspace{-30pt}  
\end{wrapfigure}

We first discuss the most popular paradigm: \emph{parameter-based} updates. 
A common view is that their gains come mainly from optimizing parameters more effectively, e.g., taking stronger updates or designing better losses. 
Our controlled analysis suggests a different view: the gradient update is only one part of the pipeline, and its effect is largely determined by factors such as available test-time evidence and a reliable unsupervised proxy. 
We select two representative methods TPT and TPS in this section.

\begin{table*}[t]
\centering
\small
\caption{\textbf{Main Results on Natural and Corruption Shifts.} Accuracy (\%) is reported for CLIP ViT-B/16. Values in brackets ($\downarrow$) denote the performance drop relative to the original ImageNet. Best and second-best results are \textbf{bolded} and \underline{underlined}, respectively.}
\vspace{-8pt}
\label{tab:main-results}
\begin{adjustbox}{width=0.82\textwidth}
\begin{tabular}{ll c | cccc c | c}
\toprule
\multirow{2}{*}{\textbf{Paradigm}} & \multirow{2}{*}{\textbf{Method}} & \textbf{Original} & \multicolumn{5}{c|}{\textbf{Natural Shifts}} & \textbf{Corruption} \\
\cmidrule(lr){3-3} \cmidrule(lr){4-8} \cmidrule(lr){9-9}
& & \textbf{INet} & \textbf{I-V2} & \textbf{I-A} & \textbf{I-R} & \textbf{I-S} & \textbf{Avg.} & \textbf{Avg.} \\
\midrule
\multicolumn{2}{l}{Zero-Shot CLIP (ViT-B/16)} & 66.72 & 60.86 & 47.87 & 73.98 & 46.08 & 57.20 \scriptsize{(\textcolor{collapse_red}{$\downarrow$9.52})} & 25.51 \scriptsize{(\textcolor{collapse_red}{$\downarrow$41.21})} \\
\midrule

% --- Parameter-based (Orange) ---
\multirow{8}{*}{\textcolor{param_orange}{\emph{Parameter-based}}} 

& TPT         & 68.85 & 63.06 & 52.12 & 76.52 & 47.94 & 59.91 \scriptsize{(\textcolor{collapse_red}{$\downarrow$8.94})} & 25.55 \scriptsize{(\textcolor{collapse_red}{$\downarrow$43.30})} \\
& DiffTPT     & 68.17 & 61.93 & 50.87 & 76.33 & 47.36 & 59.12 \scriptsize{(\textcolor{collapse_red}{$\downarrow$9.05})} & 26.07 \scriptsize{(\textcolor{collapse_red}{$\downarrow$42.10})} \\
& HisTPT         & 66.60 & 60.84 & 47.79 & 73.94 & 45.71 & 57.07 \scriptsize{(\textcolor{collapse_red}{$\downarrow$9.53})} & 25.38 \scriptsize{(\textcolor{collapse_red}{$\downarrow$41.22})} \\
& C-TPT         & 65.91 & 60.04 & 46.81 & 71.66 & 43.97 & 55.62 \scriptsize{(\textcolor{collapse_red}{$\downarrow$10.29})} & 24.80 \scriptsize{(\textcolor{collapse_red}{$\downarrow$41.11})} \\
& A-TPT         & 68.46 & 62.67 & 51.69 & 76.98 & 47.53 & 59.72 \scriptsize{(\textcolor{collapse_red}{$\downarrow$8.74})} & 25.19 \scriptsize{(\textcolor{collapse_red}{$\downarrow$43.27})} \\
& R-TPT         & 68.68 & 62.80 & 53.11 & 76.20 & 46.40 & 59.63 \scriptsize{(\textcolor{collapse_red}{$\downarrow$9.05})} &4.54 \scriptsize{(\textcolor{collapse_red}{$\downarrow$64.14})} \\

& TPS         & 70.29 & \textbf{64.01} & \textbf{58.44} & \textbf{79.30} & 50.28 & \textbf{63.01} \scriptsize{(\textcolor{collapse_red}{$\downarrow$7.28})} & 27.43 \scriptsize{(\textcolor{collapse_red}{$\downarrow$42.86})} \\
& DPE         & 70.20 & 63.61 & \underline{56.92} & 77.66 & 50.79 & 62.25 \scriptsize{(\textcolor{collapse_red}{$\downarrow$7.96})} & 27.74 \scriptsize{(\textcolor{collapse_red}{$\downarrow$42.46})} \\
& PromptAlign & 69.69 & 63.16 & 53.53 & 78.46 & 48.50 & 60.91 \scriptsize{(\textcolor{collapse_red}{$\downarrow$8.78})} & 28.75 \scriptsize{(\textcolor{collapse_red}{$\downarrow$40.94})} \\
& RLCF         & 67.43 & 61.48 & 49.11 & 74.99 & 46.87 & 58.11 \scriptsize{(\textcolor{collapse_red}{$\downarrow$9.32})} & 25.89 \scriptsize{(\textcolor{collapse_red}{$\downarrow$41.54})} \\

% & Tent        & 67.04 & 60.69 & 47.93 & 74.85 & 46.74 & 57.55 \scriptsize{(\textcolor{collapse_red}{$\downarrow$9.49})} & 26.94 \scriptsize{(\textcolor{collapse_red}{$\downarrow$40.10})} \\
% & SAR         & 66.61 & 60.77 & 48.08 & 74.56 & 46.46 & 57.47 \scriptsize{(\textcolor{collapse_red}{$\downarrow$9.14})} & \underline{31.62} \scriptsize{(\textcolor{collapse_red}{$\downarrow$34.99})} \\
% & DeYO         & 66.13 & 48.41 & 60.69 & 75.27 & 47.07 & 57.86 \scriptsize{(\textcolor{collapse_red}{$\downarrow$8.27})} & \textbf{34.98} \scriptsize{(\textcolor{collapse_red}{$\downarrow$31.15})} \\
& BATCLIP     & 66.92 & 60.94 & 48.57 & 75.99 & 44.33 & 57.46 \scriptsize{(\textcolor{collapse_red}{$\downarrow$9.46})} & \textbf{30.91} \scriptsize{(\textcolor{collapse_red}{$\downarrow$36.01})} \\
\midrule

% --- State-based (Green) ---
\multirow{5}{*}{\textcolor{state_green}{\emph{State-based}}}
& TDA         & 68.43 & 61.52 & 48.83 & 75.06 & 48.57 & 58.50 \scriptsize{(\textcolor{collapse_red}{$\downarrow$9.94})} & 27.86 \scriptsize{(\textcolor{collapse_red}{$\downarrow$40.57})} \\
& BoostAdapter& 68.39 & 61.41 & 48.96 & 75.42 & 48.90 & 58.67 \scriptsize{(\textcolor{collapse_red}{$\downarrow$9.72})} & 28.03 \scriptsize{(\textcolor{collapse_red}{$\downarrow$40.36})} \\
& DMN         & 67.54 & 61.13 & 48.40 & 74.53 & 47.17 & 57.81 \scriptsize{(\textcolor{collapse_red}{$\downarrow$9.73})} & 25.80 \scriptsize{(\textcolor{collapse_red}{$\downarrow$41.74})} \\
& ECALP       & 70.18 & 62.27 & 49.12 & 77.69 & \underline{51.74} & 60.21 \scriptsize{(\textcolor{collapse_red}{$\downarrow$9.97})} & \underline{29.76} \scriptsize{(\textcolor{collapse_red}{$\downarrow$40.42})} \\
& DoTA        & \underline{70.70} & \underline{63.74} & 58.43 & 77.66 & 50.27 & \underline{62.53} \scriptsize{(\textcolor{collapse_red}{$\downarrow$8.18})} & 25.81 \scriptsize{(\textcolor{collapse_red}{$\downarrow$44.89})} \\
& OGA         & 68.36 & 60.64 & 48.75 & 75.81 & 49.38 & 58.65 \scriptsize{(\textcolor{collapse_red}{$\downarrow$9.72})} & 27.37 \scriptsize{(\textcolor{collapse_red}{$\downarrow$40.99})} \\
& BCA         & 68.14 & 63.08 & 56.08 & 75.25 & 35.81 & 57.56 \scriptsize{(\textcolor{collapse_red}{$\downarrow$10.59})} & 3.93 \scriptsize{(\textcolor{collapse_red}{$\downarrow$64.21})} \\
& OnZeta      & \textbf{71.12} & 63.34 & 48.55 & 75.48 & \textbf{52.06} & 59.86 \scriptsize{(\textcolor{collapse_red}{$\downarrow$11.26})} & 29.38 \scriptsize{(\textcolor{collapse_red}{$\downarrow$41.74})} \\
\midrule

% --- Inference-based (Blue) ---
\multirow{4}{*}{\textcolor{infer_blue}{\emph{Inference-based}}} 
& ZERO        & 68.94 & 63.26 & 56.08 & 76.67 & 47.72 & 60.93 \scriptsize{(\textcolor{collapse_red}{$\downarrow$8.01})} & 22.95 \scriptsize{(\textcolor{collapse_red}{$\downarrow$45.99})} \\
& MTA         & 69.02 & 63.17 & 54.65 & 76.88 & 48.36 & 60.77 \scriptsize{(\textcolor{collapse_red}{$\downarrow$8.25})} & 25.42 \scriptsize{(\textcolor{collapse_red}{$\downarrow$43.60})} \\

& Panda    & 65.86 & 60.12 & 48.28 & 74.22 & 46.59 & 57.30 \scriptsize{(\textcolor{collapse_red}{$\downarrow$8.56})} & 26.89 \scriptsize{(\textcolor{collapse_red}{$\downarrow$38.97})} \\
& CALIP       & 68.39 & 61.88 & 50.03 & \underline{77.72} & 48.24 & 59.47 \scriptsize{(\textcolor{collapse_red}{$\downarrow$8.92})} & 27.13 \scriptsize{(\textcolor{collapse_red}{$\downarrow$41.26})} \\
\bottomrule
\end{tabular}
\end{adjustbox}
\vspace{-16pt}
\end{table*}

% \paragraph{Experimental setup.}

\paragraph{Stronger updates bring limited returns.}
\label{sec:tricks:steps}
The most natural explanation for parameter-based gains is that stronger test-time updates allow the model to fit each sample more precisely. 
In this view, we choose the learning rate and adaptation steps as two forms of update control for controlled experiments. 
The former controls the magnitude of each gradient update, while the latter controls the depth of adaptation for each sample.
Fig.~\ref{fig:lr_step}(a) shows the performance under different learning rates. 
Unlike the severe collapse often observed in classical TTA, TTA4CLIP methods are overall more robust to learning-rate changes. 
% We believe this might stem from fact that these methods avoid optimizing norm layers, which are highly sensitive to distributional statistics. 
However, the accuracy still fluctuates substantially, leading to potentially suboptimal performance. 
For example, the performance gap between the best and the worst choice reaches $5.3\%$ for TPT and even $7.1\%$ for TPS. 
Since it is unrealistic to perform task-specific hyperparameter tuning, expecting a universally optimal learning rate is impractical.
Fig.~\ref{fig:lr_step}(b) sweeps the number of adaptation steps from the default value of $1$ to $20$, and records both the performance gain over the previous recorded step and overall wall-clock time. 
We can see that TPT climbs from $52.0\%$ at step~$1$ to only $56.1\%$ at step~$20$, a nearly $4\%$ gain that comes at a $15\times$ cost (1,546\,s~$\to$~23,943\,s). 
App.~\ref{app:exp_hyperparameters} further shows that TPS is effectively saturated after the first step, achieving less than $0.8\%$ gains thereafter. 
The gains from additional steps quickly become marginal, while compute scales almost linearly, forming a typical diminishing-returns pattern. 
% Overall, stronger updates do bring improvements, but the gains are limited. 
% This motivates us to explore more determining drivers in test-time parameter update.
Overall, while stronger updates improve performance, the gains are limited and diminishing, motivating us to explore key drivers of test-time updates.

\begin{figure}[htbp]
    \centering
    \vspace{-10pt}
    \includegraphics[width=0.99\linewidth]{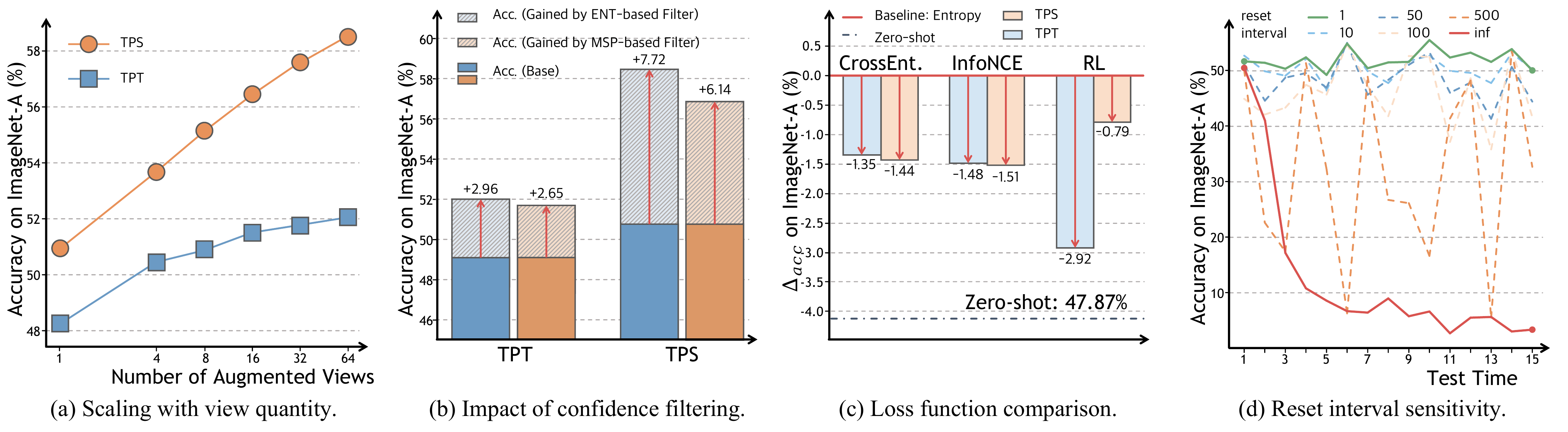}
    \vspace{-8pt}
    \caption{\textbf{Deconstructing the determining factors of test-time parameter update.} First, performance gain is largely driven by (a) the quantity and (b) the quality of test-time evidence. Furthermore, (c) the specific form of unsupervised proxy is secondary to its alignment with predictive correctness, and (d) online parameter updates collapse without stability controls such as periodic resets.}
    \label{fig:tpt_exp}
    \vspace{-13pt}
\end{figure}

\paragraph{Test-time evidence dominates optimization.}
\label{sec:tricks:evidence}

If stronger updates are not the determining driver, what is? 
Our focus first turns to \emph{test-time evidence}, i.e., available information from the test stream. 
A common procedure in previous works constructs an augmented view for each single sample and then applies an entropy-based confidence filter to obtain a high-quality subset for entropy minimization. 
% Formally, given $N$ augmented views, only the top $\max(1, \lfloor \rho N \rfloor)$ confident views are retained. 
Following this, we ablate the number of augmented views $N$ and the filter ratio $\rho$ separately, while fixing the other, to study the quantity and quality of available evidence. 
As shown in Fig.~\ref{fig:tpt_exp}(a), we surprisingly find that increasing $N$ brings substantial and near-monotonic performance gains: 
with only one update step, TPT and TPS obtain $3.8\%$ and $7.5\%$ gains respectively, even exceeding the optimization gains in Fig.~\ref{fig:lr_step}. 
Moreover, Fig.~\ref{fig:tpt_exp}(b) simply removes the confidence filter and using all augmented views for adaptation, with App.~\ref{app:exp_confidence} further providing a controlled sweep of $\rho$. 
The results indicate that the confidence filter brings roughly $3.0\%$ and $7.8\%$ gains for TPT and TPS, also comparable to the gains from evidence quantity. 
Together, test-time evidence, with a sufficient quantity and quality, dominates the gains in parameter-based TTA. 
As we show later, such gain can remain substantial even without parameter updates. 
An auxiliary analysis in App.~\ref{app:exp_augmix} further shows that the widely adopted AugMix~\cite{augmix} augmentation is not always superior to simpler random crops, leading to an underexplored, data-centric evidence design as a potential future direction. 
\begin{wrapfigure}{r}{0.5\linewidth}
    \vspace{-10pt}  
    \centering
    \includegraphics[width=0.99\linewidth]{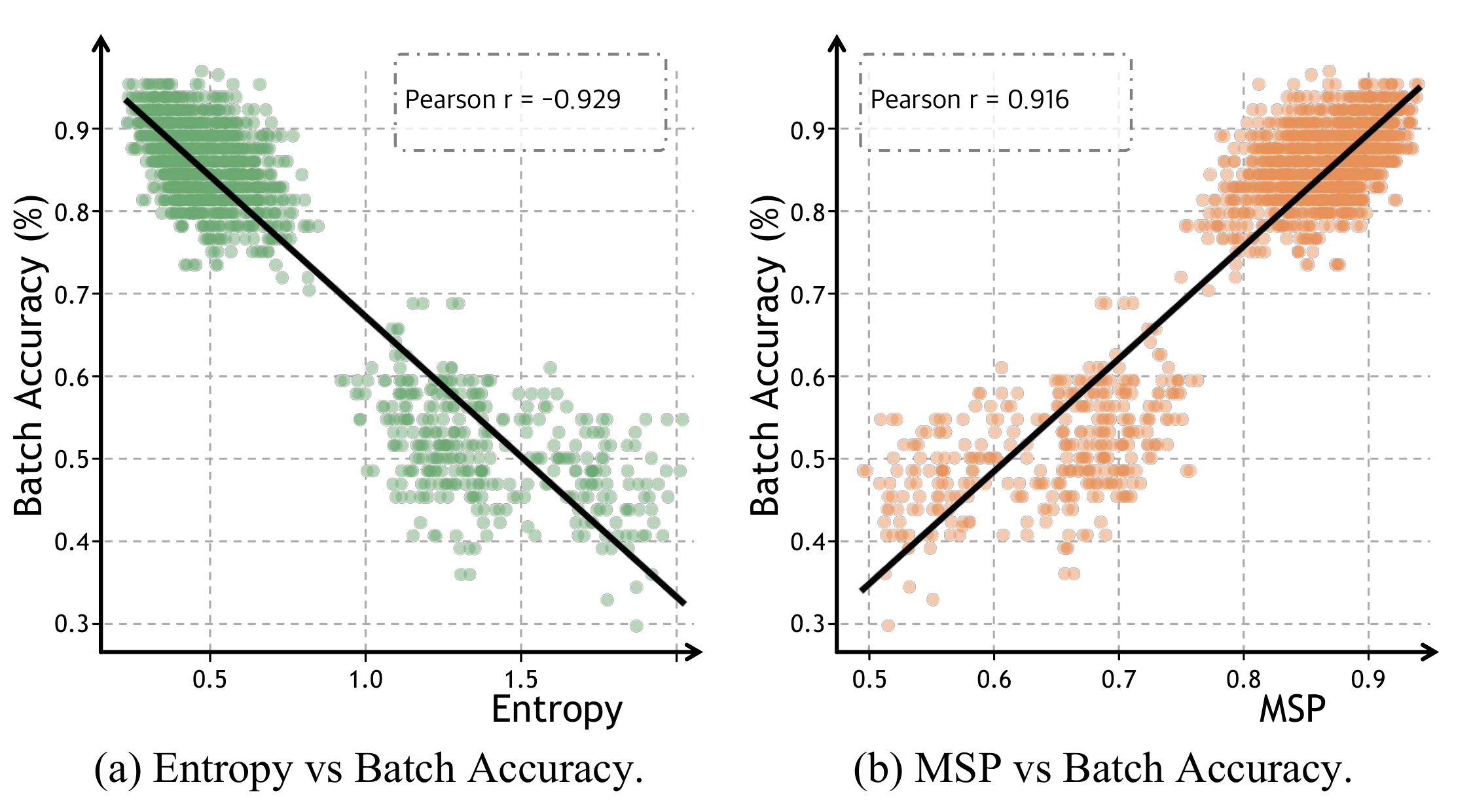}
    \vspace{-18pt}
    \captionsetup{width=0.99\linewidth}  % Set caption width to match image width
    \caption{\textbf{Correlation between Unsupervised Proxies and Batch Accuracy.} Both (a) entropy and (b) MSP exhibit a strong Pearson correlation with the ground-truth batch accuracy.}
    \label{fig:pearson}
    \vspace{-25pt}  
\end{wrapfigure}

\paragraph{Reliable proxies enable effective self-supervised update.}
\label{sec:tricks:proxy}

% The previous subsection shows that test-time evidence counts, as long as it is sufficiently broad and clean. 
The next question is: how should such evidence be used when labels are unavailable?
Parameter-based TTA requires an unsupervised signal that can decide which views are reliable and in which direction the gradient should move. 
We refer to such signals as \emph{test-time proxies}. 
A proxy may be used as a self-supervised learning objective, such as entropy minimization or contrastive learning, or as a score for evidence selection, such as confidence filtering. 
% Although these two uses are different, they serve the same purpose: converting unlabeled evidence into useful adaptation signals. 
Fig.~\ref{fig:tpt_exp}(c) compares several learning objectives under the same evidence budget, including pseudo-label cross-entropy, InfoNCE~\cite{infonce}, and CLIP-reward test-time RL~\cite{rlcf}. 
Among these objectives, entropy minimization remains the strongest and most stable proxy choice in our experiments. 
Nevertheless, alternative objectives still provide consistent performance gains. 
This suggests that proxy effectiveness may come from providing a reliable signal, rather than from its specific functional form.
We further examine this interpretation through a statistical analysis. 
In addition to entropy, we consider maximum softmax probability (MSP), defined as the largest predicted class probability. 
As shown in Fig.~\ref{fig:pearson}, entropy and MSP both exhibit strong correlation with ground-truth accuracy, with Pearson correlation coefficients of $-0.929$ and $0.916$, respectively. 
We then use MSP as an alternative evidence-selection score in Fig.~\ref{fig:tpt_exp}(b). 
As demonstrated, MSP performs similarly to entropy, effectively filtering reliable views and improving performance. 
Overall, the exact form of the proxy is secondary, as long as it is sufficiently aligned with predictive correctness. 
A useful proxy does not have to be the optimization loss itself; it can serve either as a learning objective or as a filtering criterion. 
What matters is whether it can identify reliable evidence or provide a useful update direction from unlabeled test samples. 
In this view, future works should consider designing more reliable, correctness-aligned proxies to extract useful signals from test-time evidence.

\section{Evidence Utilization Beyond Heavy Optimization}
\label{sec:metrics}
\begin{wrapfigure}{l}{0.4\textwidth}
    \centering
    \small
    \vspace{-20pt}
    \captionof{table}{\textbf{Memory Efficiency Comparison.} We record peak memory usage in GiB. Symbols ($\uparrow/\downarrow$) denote the relative change compared to zero-shot CLIP.}
    \vspace{-6pt}
    \label{tab:memory}
    \resizebox{0.90\linewidth}{!}{
    \begin{tabular}{ll c}
        \toprule
        \textbf{Paradigm} & \textbf{Method} & \textbf{Mem. (GiB)} \\
        \midrule
        \multicolumn{2}{l}{Zero-shot} & 0.75 \\
        \midrule
        \multirow{3}{*}{\textcolor{param_orange}{\emph{Param.}}} 
        & TPS         & \textbf{0.94} \scriptsize{(\textcolor{collapse_red}{$\uparrow$0.19})} \\
        & TPT         & 3.80 \scriptsize{(\textcolor{collapse_red}{$\uparrow$3.05})} \\
        & BATCLIP     & 8.54 \scriptsize{(\textcolor{collapse_red}{$\uparrow$7.79})} \\
        \midrule
        \multirow{3}{*}{\textcolor{state_green}{\emph{State}}} 
        & OnZeta      & \underline{1.31} \scriptsize{(\textcolor{collapse_red}{$\uparrow$0.56})} \\
        & TDA         & 1.34 \scriptsize{(\textcolor{collapse_red}{$\uparrow$0.59})} \\
        & DoTA        & 1.99 \scriptsize{(\textcolor{collapse_red}{$\uparrow$1.24})} \\
        \midrule
        \multirow{3}{*}{\textcolor{infer_blue}{\emph{Infer.}}} 
        & ZERO        & 1.78 \scriptsize{(\textcolor{collapse_red}{$\uparrow$1.03})} \\
        & MTA         & 1.65 \scriptsize{(\textcolor{collapse_red}{$\uparrow$0.90})} \\
        & Panda    & 2.02 \scriptsize{(\textcolor{collapse_red}{$\uparrow$1.27})} \\
        \bottomrule
    \end{tabular}
    }
    \vspace{-40pt}
\end{wrapfigure}

Sec.~\ref{sec:tricks} shows that parameter optimization contributes far less than prior literature states.
We now step beyond this paradigm and move to a popular alternative: how can CLIP exploit test-time evidence without relying on heavy back-propagation?
Our results suggest that it can be used in multiple ways: both exploiting current or historical evidence to refine predictions in a training-free manner and confining parameter updates to prototype-level residuals are effective.
However, once parameter updates accumulate online, both efficiency and stability become concerns.

% \paragraph{Experimental setup.}
\paragraph{Evidence can be exploited without optimization.}
\label{sec:metrics:history}
Tab.~\ref{tab:main-results} and ~\ref{tab:fine_grained_results} report the performance of several TTA4CLIP methods on natural-shift and fine-grained datasets, grouped by paradigm. 
For natural shifts, the best state-based method DoTA reaches $64.16\%$, nearly matching the best parameter-based method TPS ($64.46\%$), and clearly surpassing common prompt-tuning baselines such as TPT ($61.70\%$) and DiffTPT ($60.93\%$). 
As for fine-grained classification, the contrast becomes stronger: top five methods are all state-based, including OnZeta, ECALP, BoostAdapter, TDA, and OGA, ahead of every parameter-based method. 
Inference-based methods such as ZERO and CALIP sit in the middle of both rankings, despite using neither gradients nor external state. 
\begin{table*}[t]
\centering
\scriptsize 
\caption{\textbf{Main Results on Fine-grained Shifts.} Accuracy (\%) is reported for CLIP ViT-B/16 across 11 downstream datasets. Values in brackets ($\downarrow$) denote the performance drop relative to the original ImageNet. Best and second-best results are \textbf{bolded} and \underline{underlined}, respectively.}
\vspace{-8pt}
\label{tab:fine_grained_results}
\begin{adjustbox}{width=\textwidth}
\begin{tabular}{ll c | cccccccccc | c}
\toprule
\multirow{2}{*}{\textbf{Paradigm}} & \multirow{2}{*}{\textbf{Method}} & \textbf{Original} & \multicolumn{10}{c|}{\textbf{Fine-grained Datasets}} & \textbf{Overall} \\
\cmidrule(lr){3-3} \cmidrule(lr){4-13} \cmidrule(lr){14-14}
& & \textbf{INet} & \textbf{Cal} & \textbf{Pets} & \textbf{Cars} & \textbf{FLW} & \textbf{Food} & \textbf{Air} & \textbf{SUN} & \textbf{DTD} & \textbf{SAT} & \textbf{UCF} & \textbf{Avg.} ($\downarrow$) \\
\midrule
\multicolumn{2}{l}{Zero-Shot} & 66.72 & 93.31 & 88.20 & 65.51 & 67.48 & 85.22 & 23.76 & 62.55 & 44.39 & 42.02 & 65.11 & 63.76 \scriptsize{(\textcolor{collapse_red}{$\downarrow$2.96})} \\
\midrule

% --- Parameter-based (Orange) ---
\multirow{9}{*}{\textcolor{param_orange}{\emph{Parameter-based}}} 

& TPT          & 68.85 & 93.47 & 88.17 & 67.04 & 68.94 & 86.29 & 23.73 & 65.50 & 46.57 & 42.04 & 68.17 & 64.99 \scriptsize{(\textcolor{collapse_red}{$\downarrow$3.86})} \\
& DiffTPT      & 68.17 & 93.55 & 88.55 & 65.86 & 68.62 & 86.04 & 23.97 & 65.29 & 46.22 & 44.54 & 68.15 & 65.08 \scriptsize{(\textcolor{collapse_red}{$\downarrow$3.09})} \\
& HisTPT       & 66.60 & 93.18 & 88.28 & 65.44 & 69.55 & 85.25 & 23.85 & 62.42 & 44.33 & 41.72 & 65.21 & 63.92 \scriptsize{(\textcolor{collapse_red}{$\downarrow$2.68})} \\
& C-TPT          & 65.91 & 91.6 & 86.21 & 65.09 & 70.89 & 83.95 & 23.46 & 62.89 & 43.03 & 37.32 & 62.25 & 62.67 \scriptsize{(\textcolor{collapse_red}{$\downarrow$3.24})} \\
& A-TPT        & 68.46 & 91.85 & 84.08 & 63.28 & 66.06 & 84.67 & 23.61 & 65.79 & 45.45 & 44.30 & 62.52 & 63.16 \scriptsize{(\textcolor{collapse_red}{$\downarrow$5.30})} \\
& R-TPT         & 68.68 & 93.79 & 86.94 & 67.16 & 68.33 & 85.96 & 23.73 & 65.58 & 46.04 & 34.67 & 67.35 & 63.95 \scriptsize{(\textcolor{collapse_red}{$\downarrow$4.73})} \\
& TPS          & 70.29 & 93.83 & 80.68 & 68.54 & \textbf{71.70} & 82.93 & 25.77 & 63.70 & \textbf{54.31} & 42.64 & 67.59 & 65.17 \scriptsize{(\textcolor{collapse_red}{$\downarrow$5.12})} \\
& DPE          & 70.20 & \textbf{95.13} & 89.15 & 69.43 & 70.28 & 86.23 & 25.47 & 68.44 & 49.23 & 37.06 & 70.42 & 66.08 \scriptsize{(\textcolor{collapse_red}{$\downarrow$4.12})} \\
& PromptAlign  & 69.69 & 93.10 & 85.77 & 67.19 & 66.06 & 86.65 & 23.82 & 65.90 & 43.79 & 47.20 & 67.54 & 64.70 \scriptsize{(\textcolor{collapse_red}{$\downarrow$4.99})} \\
& RLCF      & 67.43 & 93.63 & 88.77 & 66.22 & 69.18 & 85.59 & 24.30 & 63.51 & 45.57 & 42.38 & 66.72 & 64.59 \scriptsize{(\textcolor{collapse_red}{$\downarrow$2.84})} \\
% & Tent      & 67.04 & 93.55 & 88.33 & 65.64 & 67.38 & 85.18 & 23.88 & 62.91 & 44.50 & 46.28 & 65.45 & 64.29 \scriptsize{(\textcolor{collapse_red}{$\downarrow$2.75})} \\
% & SAR      & 66.61 & 93.31 & 88.28 & 65.5 & 67.52 & 85.21 & 23.85 & 62.53 & 44.33 & 43.58 & 65.21 & 63.93 \scriptsize{(\textcolor{collapse_red}{$\downarrow$2.68})} \\
% & DeYO      & 66.13 & 93.14 & 88.55 & 65.69 & 67.19 & 85.16 & 23.49 & 61.65 & 44.56 & 47.68 & 66.22 & 64.33 \scriptsize{(\textcolor{collapse_red}{$\downarrow$1.80})} \\
& BATCLIP      & 66.92 & 93.55 & 89.04 & 64.93 & 68.53 & 85.26 & 23.28 & 65.73 & 44.68 & 40.21 & 65.82 & 64.10 \scriptsize{(\textcolor{collapse_red}{$\downarrow$2.82})} \\
\midrule

% --- State-based (Green) ---
\multirow{5}{*}{\textcolor{state_green}{\emph{State-based}}} 

& TDA          & 68.43 & 93.67 & 89.21 & 66.76 & 70.28 & 85.64 & 23.64 & 65.72 & 47.34 & 54.14 & 68.52 & 66.49 \scriptsize{(\textcolor{collapse_red}{$\downarrow$1.94})} \\
& BoostAdapter & 68.39 & 93.79 & 89.18 & 66.53 & 70.32 & 85.82 & 23.70 & 65.81 & 46.87 & 54.68 & 68.75 & 66.55 \scriptsize{(\textcolor{collapse_red}{$\downarrow$1.84})} \\
& DMN & 67.54 & 93.43 & 88.50 & 66.14 & 68.05 & 85.57 & 23.67 & 63.97 & 45.51 & 46.36 & 66.85 & 64.81 \scriptsize{(\textcolor{collapse_red}{$\downarrow$2.74})} \\
& ECALP        & 70.18 & 93.27 & 89.15 & 68.47 & 71.13 & \underline{87.14} & \underline{26.04} & \underline{68.94} & 47.81 & \underline{55.19} & \underline{73.17} & \underline{68.03} \scriptsize{(\textcolor{collapse_red}{$\downarrow$2.15})} \\
& DoTA       & \underline{70.70} & \underline{94.44} & 88.42 & \underline{70.40} & 68.98 & 85.13 & 25.53 & 68.83 & 48.94 & 39.37 & 68.97 & 65.90 \scriptsize{(\textcolor{collapse_red}{$\downarrow$4.80})} \\
& OGA          & 68.36 & 92.94 & \underline{89.70} & 67.75 & 70.08 & 85.58 & 23.76 & 66.24 & 47.64 & 53.90 & 69.36 & 66.70 \scriptsize{(\textcolor{collapse_red}{$\downarrow$1.66})} \\
& BCA & 68.14 & 94.08 & 85.83 & 67.27 & 66.38 & 83.49 & 23.19 & 64.65 & 45.15 & 35.52 & 66.19 & 63.18 \scriptsize{(\textcolor{collapse_red}{$\downarrow$4.96})} \\
& OnZeta       & \textbf{71.12} & 91.12 & \textbf{91.77} & \textbf{70.68} & \underline{71.17} & \textbf{87.59} & \textbf{27.33} & \textbf{70.43} & \underline{50.30} & \textbf{57.95} & \textbf{75.55} & \textbf{69.39} \scriptsize{(\textcolor{collapse_red}{$\downarrow$1.73})} \\
\midrule

% --- Inference-based (Blue) ---
\multirow{4}{*}{\textcolor{infer_blue}{\emph{Inference-based}}} 
& ZERO         & 68.94 & 94.04 & 87.30 & 67.28 & 66.71 & 85.38 & 25.44 & 65.48 & 45.74 & 37.20 & 66.51 & 64.11 \scriptsize{(\textcolor{collapse_red}{$\downarrow$4.83})} \\
& MTA          & 69.02 & 94.04 & 87.95 & 67.59 & 67.40 & 86.07 & 24.39 & 65.28 & 45.80 & 42.49 & 67.62 & 64.86 \scriptsize{(\textcolor{collapse_red}{$\downarrow$4.16})} \\

& Panda     & 65.86 & 91.68 & 88.06 & 65.15 & 66.59 & 84.91 & 23.52 & 62.70 & 44.39 & 48.90 & 65.87 & 64.18 \scriptsize{(\textcolor{collapse_red}{$\downarrow$1.68})} \\
& CALIP        & 68.39 & 93.63 & 88.31 & 66.34 & 66.14 & 85.34 & 23.88 & 65.86 & 45.27 & 47.44 & 67.09 & 64.93 \scriptsize{(\textcolor{collapse_red}{$\downarrow$3.46})} \\
\bottomrule
\end{tabular}
\end{adjustbox}
\vspace{-12pt}
\end{table*}

These results show that test-time evidence does not have to be converted into gradient updates to be useful. 
Inference-based methods exploit \emph{current-sample evidence}: for example, from augmented views or intra-image contextual features (e.g., non-\texttt{[CLS]} ViT tokens). 
State-based methods instead exploit \emph{cross-sample evidence} accumulated from the test stream. 
This may take the form of cached features, historical predictions, class-conditional distributional statistics, or relational structures such as graphs. 
Even without optimization, these methods can achieve competitive performance, and often outperform most parameter-based methods. 
Together with Sec.~\ref{sec:tricks}, we can conclude that: CLIP adapts mainly by accessing more reliable test-time evidence. 
Optimization captures only one part, whereas appropriate evidence construction and utilization can extract much of the adaptation gains with lower cost.

\paragraph{Lightweight optimization provides a middle ground.}
\label{sec:metrics:sweetspot} 

The above analysis does not imply that parameter updates are useless. 
Rather, when parameter updates are needed, the key question is not simply whether to update parameters, but \emph{where} and \emph{how} they are updated. 
Tab.~\ref{tab:main-results} shows that the lightest update scope still performs excellently. 
TPS updates only a $K \times d$ matrix of per-class prototype residuals, yet achieves $58.44\%$ on ImageNet-A and a $64.46\%$ natural-shift average, outperforming all other candidates. 
DPE, which uses dual evolving prototypes across modalities, follows closely. 
Prompt tuning occupies the middle tier, while norm-layer tuning remains near the bottom on these natural-shift benchmarks, sometimes only $1-2\%$ above zero-shot. 
This does not necessarily mean that norm-layer adaptation is ineffective in general; as discussed in Sec.~\ref{sec:shifts}, such methods are primarily designed for corruptions, where adjusting feature statistics is more relevant. 
These results suggest that the most competitive parameter-based methods do not require heavy optimization. 
Instead, they update compact variables outside the backbone, such as class-wise prototype residuals. 
This provides an efficient middle ground: lightweight residual updates preserve the flexibility of gradient-based adaptation, while reducing the cost of heavy back-propagating through large parts of the model. 
Tab.~\ref{tab:memory} further supports this view: TPS requires only $1.30$ GiB memory, even lower than those state-based methods, whereas TPT and BATCLIP require $3.80$ and $8.54$ GiB, respectively. 

\paragraph{Online parameter updating requires efficiency and stability controls.}
\label{sec:metrics:cost}
Since Sec.~\ref{sec:metrics:history} shows that cross-sample evidence can be valuable, it is natural to ask whether one can simply accumulate parameter updates over the test stream. 
Our analysis suggests that this is risky. 
Parameter-based methods can suffer from severe stability issues when adapted continuously in online settings.
To examine this issue, we modify the per-sample episodic reset mechanism of TPT and introduce a periodic reset interval $\tau$. 
The model is reset every $\tau$ samples, and we report the average accuracy within each 500-sample logging window in Fig.~\ref{fig:tpt_exp}(d). 
We observe a clear collapse phenomenon: model performance gradually decreases over the test stream, and the degradation becomes more severe as the reset interval increases. 
When reset is completely removed, model performs well in the first logging window, but rapidly drops below $10\%$ within only four windows and eventually collapses to around $3\%$ near the end of adaptation. 
This failure is not tied to a specific optimization paradigm. 
Instead, it arises from the accumulation of noisy self-supervised signals. 
Once early errors are written into parameters, later updates may reinforce these errors before useful adaptation can build up.
This observation indicates that online parameter updating requires explicit stability mechanisms in practice. 
We suggest three possible strategies: 
(i) applying periodic reset with an appropriate interval; 
(ii) integrating external state into the backward pass, as in HisTPT; 
and (iii) avoiding online parameter updates and instead combining episodic adaptation with an external state, as in DPE. 

\begin{wrapfigure}{r}{0.38\linewidth}
    \vspace{-20pt}  
    \centering
    \includegraphics[width=0.99\linewidth]{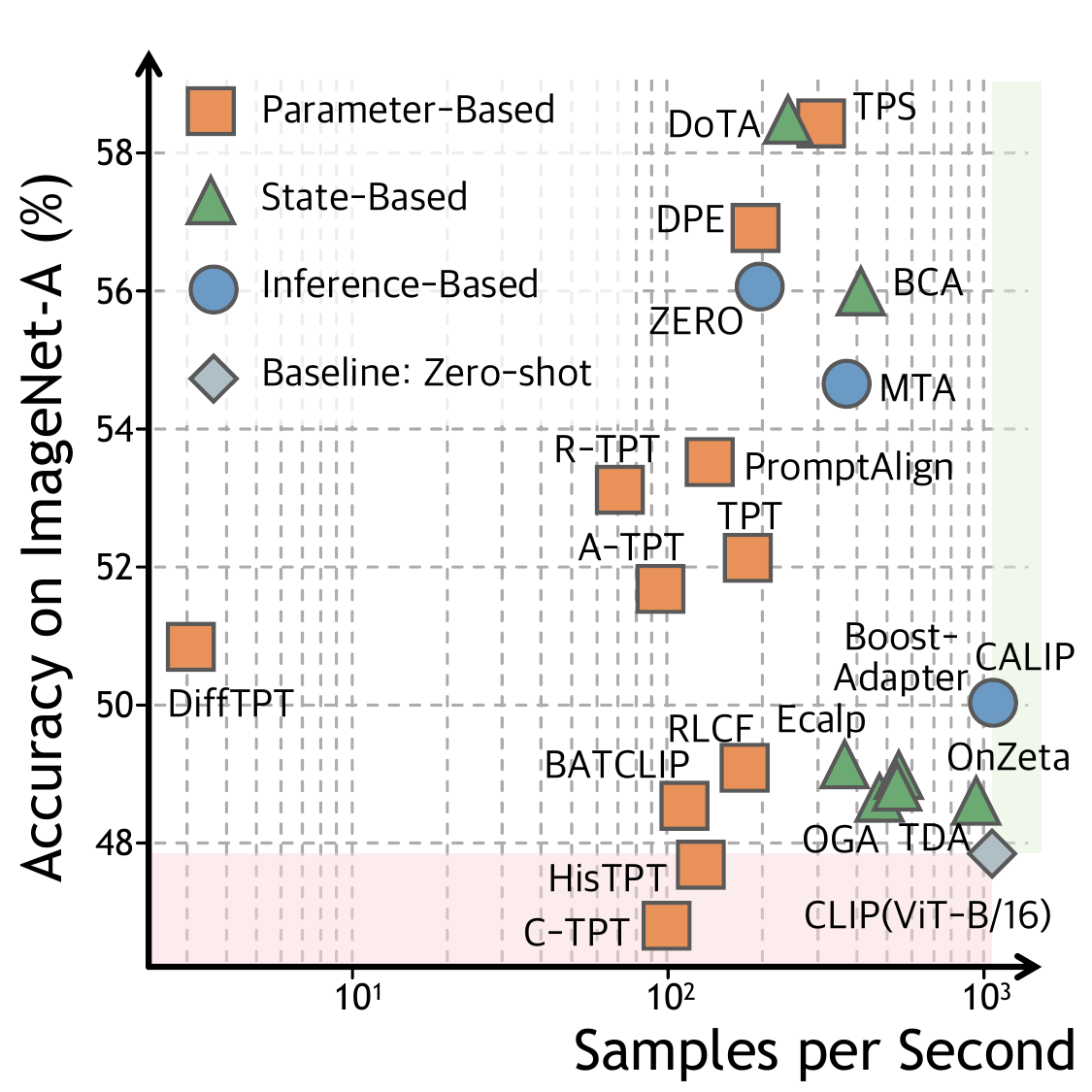}
    \captionsetup{width=0.99\linewidth}  % Set caption width to match image width
    \vspace{-18pt}
    \caption{\textbf{Comparison of accuracy and adaptation cost.} Parameter-based methods occupy the expensive region, while state/inference-based methods achieve better tradeoff.}
    % , achieving superior natural-shift accuracy with higher throughput than heavy backpropagation-based paradigms
    \label{fig:acc-cost}
    \vspace{-40pt}  
\end{wrapfigure}

Parameter updating also incurs substantial computational cost. 
Fig.~\ref{fig:acc-cost} compares the accuracy and per-sample adaptation cost (on a logarithmic scale) of all methods. 
Among them, parameter-based methods occupy the expensive region. 
DiffTPT, which relies on an external diffusion model to generate additional views, requires 21.01 seconds per sample, corresponding to nearly 44 hours on the full I-A dataset. 
TPT takes 0.35 seconds, BATCLIP takes 0.57 seconds, and even the lightweight TPS still requires 0.21 seconds.
In contrast, most state-based and inference-based methods remain around 0.10-0.15 seconds per sample and thus can finish adaptation within 20 minutes. 
% For these methods, the main overhead usually comes from additional forward passes over augmented views rather than backward propagation. 
Overall, when considering online parameter updating, practical solutions must balance accuracy with efficiency and stability.

\section{No Silver Bullet for TTA Across Shifts}
\label{sec:shifts}
Secs.~\ref{sec:tricks} and~\ref{sec:metrics} show that TTA4CLIP performance is largely determined by how each adaptation paradigm handles test-time evidence, rather than by optimization alone.
We now ask whether this update perspective remains consistent across different distribution shifts. 
The answer is complicated: the preferred adaptation paradigm changes with the nature of shift.

\paragraph{Paradigm-wise preferences vary across shifts.}
\label{sec:shifts:family}

\begin{wraptable}{l}{0.65\textwidth}
    \centering
    \small
    \vspace{-15pt}
    \caption{\textbf{Norm-layer adaptation specializes in corruption shifts.} We compare norm-layer adaptation methods (e.g., DeYO, SAR) with representative TTA4CLIP methods across diverse scenarios. 
    % While trailing in semantic shifts, these methods are indispensable for recalibrating low-level visual statistics distorted by noise or blur.
    }
    \vspace{-6pt}
    \label{tab:corr}
    \resizebox{0.99\linewidth}{!}{
    \begin{tabular}{l cccc | ccc}
        \toprule
        \multirow{2}{*}{\textbf{Method}} & \multicolumn{4}{c|}{\textbf{Characteristics}} & \multicolumn{3}{c}{\textbf{Avg. Acc (\%)}} \\
        \cmidrule(lr){2-5} \cmidrule(lr){6-8}
        & \textbf{Norm} & \textbf{Prom.} & \textbf{Filt.} & \textbf{Mult.} & \textbf{Nat.} & \textbf{Fine.} & \textbf{Corr.} \\
        \midrule
        CLIP & \ding{55} & \ding{55} & \ding{55} & \ding{55} & 57.20 & 63.76 & 25.51 \\
        \midrule
        % --- Prompt-based ---
        TPT & \ding{55} & \ding{51} & \ding{51} & \ding{55} & \underline{59.91} & 64.99 & 25.55 \\
        TPS & \ding{55} & \ding{51} & \ding{51} & \ding{55} & \textbf{63.01} & 65.17 & 27.43 \\
        TDA & \ding{55} & \ding{55} & \ding{51} & \ding{51} & 58.50 & \underline{66.49} & 27.86 \\
        OnZeta & \ding{55} & \ding{55} & \ding{55} & \ding{51} & 59.86 & \textbf{69.39} & 29.38 \\
        \midrule
        % --- Norm-tuning ---
        BATCLIP & \ding{51} & \ding{51} & \ding{55} & \ding{51} & 57.46 & 64.10 & 30.91 \\
        Tent & \ding{51} & \ding{55} & \ding{55} & \ding{55} & 57.55 & 64.29 & 26.94 \\
        SAR & \ding{51} & \ding{55} & \ding{51} & \ding{55} & 57.47 & 63.93 & 31.62 \\
        DeYO & \ding{51} & \ding{55} & \ding{51} & \ding{55} & 57.86 & 64.33 & \textbf{34.98} \\
        Tent w. Panda & \ding{51} & \ding{55} & \ding{51} & \ding{51} & 57.77 & 64.49 & \underline{31.78} \\
        \bottomrule
    \end{tabular}
    }
    \vspace{-10pt}
\end{wraptable}

As shown in Tabs.~\ref{tab:main-results} and~\ref{tab:fine_grained_results}, paradigm preferences diverge significantly between natural shifts and fine-grained classification. 
On natural shifts, lightweight parameter-based (e.g., prototype updates) and inference-based methods generally excel. 
Since these shifts preserve the original label space and merely introduce superficial visual variations, adjusting visual-textual alignment around existing, meaningful text anchors or aggregating diverse augmented views provides sufficient robustness. 

Conversely, state-based methods dominate fine-grained datasets. These tasks often introduce novel, domain-specific classes far from the training distribution. In such scenarios, single-sample augmentations cannot recover the missing class structure. State-based methods are therefore advantageous, as they accumulate cross-sample evidence to organically construct dataset-specific class distributions. 
Overall, no single paradigm is universally optimal; the key is ensuring that a method's evidence source and update mechanism align with the structural nature of the shift.

\paragraph{Corruption favors objective and evidence-aware design.}
\label{sec:shifts:corr}
We further consider image corruptions, a standard setting in classical TTA.  
In addition to existing methods in Tab.~\ref{tab:main-results}, we implement 3 representative classical TTA baselines on CLIP, including Tent~\cite{Tent}, SAR~\cite{SAR}, and DeYO~\cite{DEYO}.
% , and combine Tent with Panda.
We also combine Tent with Panda, since Panda is designed specifically for corruption. 
The results still demonstrate a different preference. 
Parameter-based norm-layer adaptation methods that are less competitive on natural and fine-grained shifts become much stronger under corruptions, while many methods that perform well on the two shifts fall to the middle. 
As shown in Tab.~\ref{tab:corr}, DeYO achieves the best accuracy of $34.98\%$, followed by Panda+Tent ($31.78\%$), SAR ($31.62\%$), and BATCLIP ($30.91\%$). 
We attribute this to the fact that corruptions, such as noise, blur, and digital artifacts directly perturb low-level visual statistics, thereby changing the activation distributions processed by the visual encoder. 
Since norm layers control how features are standardized and re-scaled, adapting them is equivalent to re-calibrating corrupted feature statistics. 
This observation is consistent with prior observation that normalization-based feature alignment is most effective for shifts in local image statistics, but offers limited gains under more abstract shifts~\cite{burns2021limitations}.

\begin{wrapfigure}{l}{0.37\linewidth}
    \vspace{-18pt}
    \centering
    \includegraphics[width=0.99\linewidth]{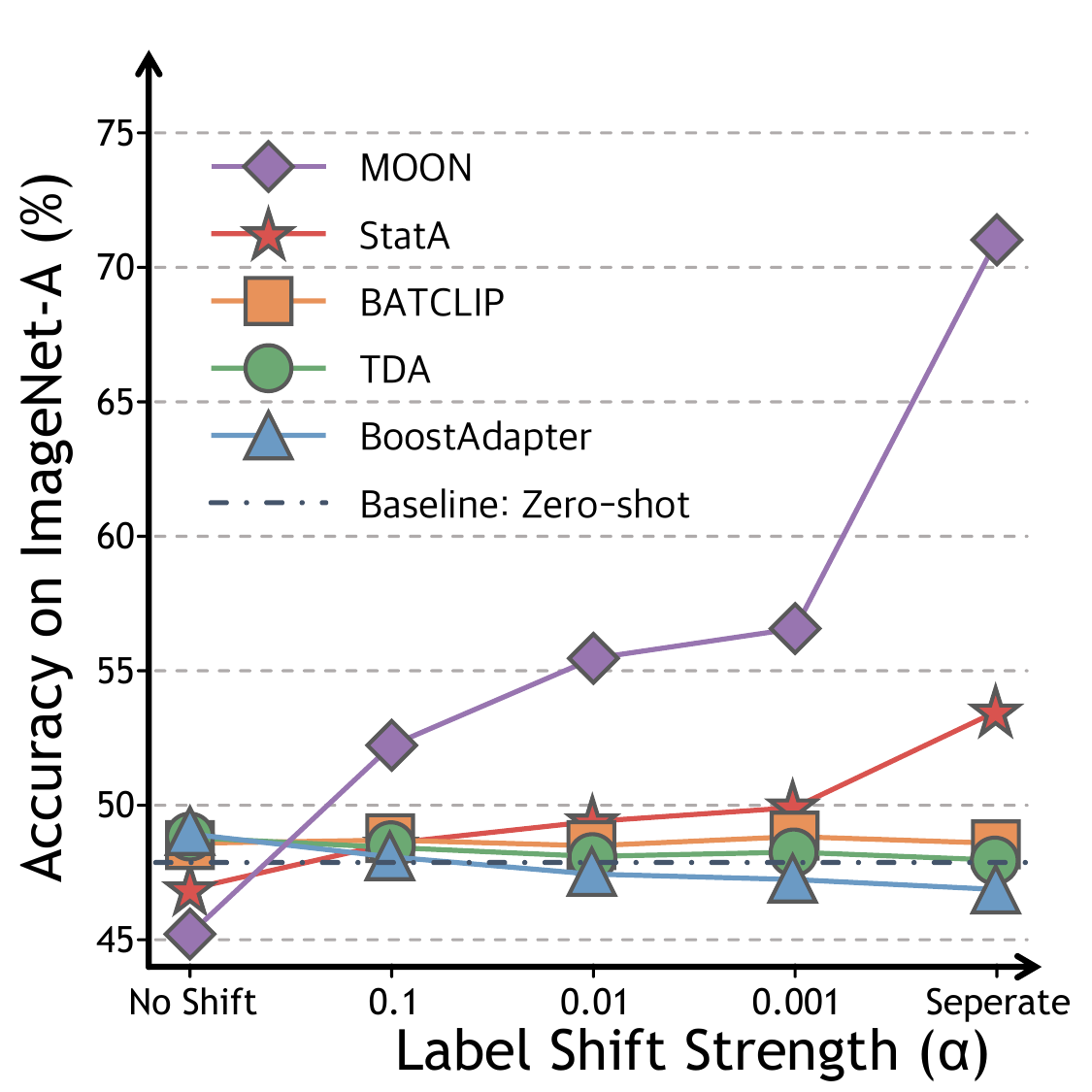}
    \vspace{-10pt}
    \captionsetup{width=0.99\linewidth}  % Set caption width to match image width
    \caption{\textbf{Performance under label shift.} Existing TTA4CLIP methods stay relatively stable, but fail to exploit the rich label priors provided. 
    % Across varying shift strengths ($\alpha$), adaptation actively hurts or merely matches the unadapted zero-shot baseline, failing to mitigate the distribution shift.
    }
    \label{fig:label-shift}
    \vspace{-18pt}  
\end{wrapfigure}

Results also show that updating more parameters is not always better. 
Starting from naive entropy minimization, designing a more reliable proxy or using more informative evidence already brings large gains. 
Compared with Tent (\(26.94\%\)), SAR outperforms by $4.68\%$ through confidence filtering and gradient reweighting, and DeYO further enlarges the gain to $8.04\%$ by exploiting evidence from local patches. 
% Similarly, combining Panda with Tent brings $4.84\%$ gains. 
By contrast, expanding the update scope to bimodal encoders, as in BATCLIP, reaches a smaller gain. 
These observations also echo the findings in Sec.~\ref{sec:tricks}. 
For corruptions, the most effective strategy is often not broad parameter updating, but an objective and evidence-aware design along with norm-layer adaptation. 

\paragraph{Label shifts call for shift-aware design.}
\label{sec:shifts:label}
Finally, we consider label shift induced by temporal correlation in the test stream. 
% This type of shift introduces temporal correlation in the test stream, and can be overlaid on any of the other three shifts.
Specifically, we construct temporally correlated streams using a Dirichlet-based protocol~\cite{dirichlet}, where smaller concentration values indicate stronger temporal correlation, and additionally consider a separate-class stream as the most extreme case. 
Details are provided in App.~\ref{app:label-shift}. 
We focus on online methods whose predictions can depend on history-dependent behavior (e.g., state-based, online parameter-based). 
Fig.~\ref{fig:label-shift} shows that existing TTA4CLIP methods are relatively stable under temporal label correlation. 
They do not exhibit the severe collapse often observed in classical online TTA, likely because CLIP retains strong open-vocabulary semantic priors and does not require adapting a closed-set classifier. 
Nevertheless, these methods still fail to exploit the rich label prior provided by the correlated stream, with most methods remaining close to zero-shot CLIP and becoming worse as the shift becomes more severe. 
To examine whether label priors can be further exploited, we include StatA~\cite{stata} and MOON~\cite{moon}, two transductive methods that specifically designed for realistic correlated streams. 
% StatA estimates the test-time class prior and uses it to calibrate predictions, allowing the model to convert temporal label structure into useful evidence. 
By performing joint inference over batches, they are able to leverage the label prior and uses it to calibrate predictions. 
% As shown, under label shift, StatA achieves consistent and significant gains, with larger improvements under stronger temporal correlation. 
As shown, when label shift becomes stronger, both methods increasingly benefit from the label structure.
In particular, MOON substantially outperforms all other methods, benefiting from its dynamic design based on both instance- and class-level label priors.
% In the extreme separate case, StatA and MOON improves from $46.89\%$ to $53.51\%$.
In the extreme separate case, StatA and MOON achieve $53.51\%$ and $71.02\%$ accuracy, respectively.
% Overall, existing TTA4CLIP methods are stable enough to avoid collapse, but they do not fully benefit from label-shifted streams. 
This highlights an open direction for future research: explicitly modeling and leveraging shifted label priors~\cite{veilleux2021realistic} as evidence. 

\section{Conclusion}
\label{sec:conclusion}

In this paper, we conduct a systematic controlled empirical study of TTA4CLIP by introducing TTABC, an open-source TTA benchmark that integrates over 20 representative methods across a diverse range of distribution shift scenarios. 
Our controlled study reveals that adaptation gains are primarily driven by test-time evidence and reliable proxies, effectively outweighing the diminishing returns of heavy parameter optimization. 
Furthermore, we demonstrate that competitive and highly efficient performance can be achieved without heavy tuning by leveraging cross- or current-sample evidence for prediction refinement, as well as lightweight prototype updates. 
Finally, we emphasize that there is no silver bullet for TTA; optimal performance requires aligning the adaptation paradigm with the specific nature of the shift (e.g., norm-layer updates for corruptions, or stream-aware designs for label shifts). 
In conclusion, TTABC establishes a solid foundation for comprehensively understanding the TTA mechanisms of CLIP and provides clear directions for developing robust, efficient, and shift-aware adaptation techniques in complex real-world environments.

\section*{Acknowledgements and Disclosures}
\noindent\textbf{Funding Statement.} This work was supported by the National Natural Science Foundation of China (Grant No. 92467204 and 62472249), the Shenzhen Science and Technology Program (Grant No. KJZD20240903102300001 and JCYJ20250604145014018), and the Natural Science Foundation of Top Talent of SZTU (Grant No. GDRC202413). 

\noindent\textbf{Competing Interests.} The authors declare no competing interests.

We thank the reviewers and area chairs for their efforts in helping us improve the quality of this paper.
% \end{ack}

% \newpage
{
% \small

\bibliographystyle{plain}  % 或者 use unsrt, ieeetr, etc.
\bibliography{mybib}

@article{TTAB,
  title={On pitfalls of test-time adaptation},
  author={Zhao, Hao and Liu, Yuejiang and Alahi, Alexandre and Lin, Tao},
  journal={arXiv preprint arXiv:2306.03536},
  year={2023}
}

@inproceedings{CLIP,
  title={Learning transferable visual models from natural language supervision},
  author={Radford, Alec and Kim, Jong Wook and Hallacy, Chris and Ramesh, Aditya and Goh, Gabriel and Agarwal, Sandhini and Sastry, Girish and Askell, Amanda and Mishkin, Pamela and Clark, Jack and others},
  booktitle={International conference on machine learning},
  pages={8748--8763},
  year={2021},
  organization={PmLR}
}

@inproceedings{resnet,
  title={Deep residual learning for image recognition},
  author={He, Kaiming and Zhang, Xiangyu and Ren, Shaoqing and Sun, Jian},
  booktitle={Proceedings of the IEEE conference on computer vision and pattern recognition},
  pages={770--778},
  year={2016}
}

@article{vit,
  title={An Image is Worth 16x16 Words: Transformers for Image Recognition at Scale},
  author={Dosovitskiy, Alexey and Beyer, Lucas and Kolesnikov, Alexander and Weissenborn, Dirk and Zhai, Xiaohua and Unterthiner, Thomas and  Dehghani, Mostafa and Minderer, Matthias and Heigold, Georg and Gelly, Sylvain and Uszkoreit, Jakob and Houlsby, Neil},
  journal={ICLR},
  year={2021}
}

@article{TTA_survey,
  title={A comprehensive survey on test-time adaptation under distribution shifts},
  author={Liang, Jian and He, Ran and Tan, Tieniu},
  journal={arXiv preprint arXiv:2303.15361},
  year={2023}
}

@article{vlmtta,
  title={The illusion of progress? a critical look at test-time adaptation for vision-language models},
  author={Sheng, Lijun and Liang, Jian and He, Ran and Wang, Zilei and Tan, Tieniu},
  journal={arXiv preprint arXiv:2506.24000},
  year={2025}
}

@article{shift_1,
  title={How transferable are features in deep neural networks?},
  author={Yosinski, Jason and Clune, Jeff and Bengio, Yoshua and Lipson, Hod},
  journal={Advances in neural information processing systems},
  volume={27},
  year={2014}
}

@article{shift_2,
  title={Right for the wrong reasons: Diagnosing syntactic heuristics in natural language inference},
  author={McCoy, R Thomas and Pavlick, Ellie and Linzen, Tal},
  journal={arXiv preprint arXiv:1902.01007},
  year={2019}
}

@inproceedings{cliff,
title = {Cliff: Leveraging Ambiguous Samples for Enhanced Test-Time Adaptation},
author = {Chen, Xiao and Zhang, Qihui and Wang, Yan},
booktitle={27th European Conference on Artificial Intelligence},
volume={392},
pages = {642-649},
year = {2024},
doi = {10.3233/FAIA240544}
}

@article{Tent,
  title={Tent: Fully test-time adaptation by entropy minimization},
  author={Wang, Dequan and Shelhamer, Evan and Liu, Shaoteng and Olshausen, Bruno and Darrell, Trevor},
  journal={arXiv preprint arXiv:2006.10726},
  year={2020}
}

@article{DEYO,
  title={Entropy is not enough for test-time adaptation: From the perspective of disentangled factors},
  author={Lee, Jonghyun and Jung, Dahuin and Lee, Saehyung and Park, Junsung and Shin, Juhyeon and Hwang, Uiwon and Yoon, Sungroh},
  journal={arXiv preprint arXiv:2403.07366},
  year={2024}
}

@inproceedings{shot, 
 title={Do We Really Need to Access the Source Data? Source Hypothesis Transfer for Unsupervised Domain Adaptation}, 
 author={Liang, Jian and Hu, Dapeng and Feng, Jiashi}, 
 booktitle={International Conference on Machine Learning (ICML)},  
 pages={6028--6039},
 year={2020}
}

@article{bn_adapt,
  author = { Schneider, Steffen and Rusak, Evgenia
    and Eck, Luisa and Bringmann, Oliver
    and Brendel, Wieland and Bethge, Matthias
  },
  title = {Removing covariate shift improves
    robustness against common corruptions
  },
  journal = {CoRR},
  volume = {abs/2006.16971},
  year = {2020},
}

@article{SAR,
  title={Towards stable test-time adaptation in dynamic wild world},
  author={Niu, Shuaicheng and Wu, Jiaxiang and Zhang, Yifan and Wen, Zhiquan and Chen, Yaofo and Zhao, Peilin and Tan, Mingkui},
  journal={arXiv preprint arXiv:2302.12400},
  year={2023}
}

@inproceedings{nctta,
  title={Neural Collapse in Test-Time Adaptation},
  author={Chen, Xiao and Du, Zhongjing and Huang, Jiazhen and Jiang, Xu and Lu, Li and Jiang, Jingyan and Wang, Zhi},
  booktitle={Proceedings of the IEEE/CVF Conference on Computer Vision and Pattern Recognition},
  pages={10567--10576},
  year={2026}
}

@article{coop,
  title={Learning to prompt for vision-language models},
  author={Zhou, Kaiyang and Yang, Jingkang and Loy, Chen Change and Liu, Ziwei},
  journal={International journal of computer vision},
  volume={130},
  number={9},
  pages={2337--2348},
  year={2022},
  publisher={Springer}
}

@inproceedings{cocoop,
  title={Conditional prompt learning for vision-language models},
  author={Zhou, Kaiyang and Yang, Jingkang and Loy, Chen Change and Liu, Ziwei},
  booktitle={Proceedings of the IEEE/CVF conference on computer vision and pattern recognition},
  pages={16816--16825},
  year={2022}
}

@inproceedings{maple,
  title={Maple: Multi-modal prompt learning},
  author={Khattak, Muhammad Uzair and Rasheed, Hanoona and Maaz, Muhammad and Khan, Salman and Khan, Fahad Shahbaz},
  booktitle={Proceedings of the IEEE/CVF conference on computer vision and pattern recognition},
  pages={19113--19122},
  year={2023}
}

@article{tpt,
  title={Test-time prompt tuning for zero-shot generalization in vision-language models},
  author={Shu, Manli and Nie, Weili and Huang, De-An and Yu, Zhiding and Goldstein, Tom and Anandkumar, Anima and Xiao, Chaowei},
  journal={Advances in Neural Information Processing Systems},
  volume={35},
  pages={14274--14289},
  year={2022}
}

@inproceedings{difftpt,
  title={Diverse data augmentation with diffusions for effective test-time prompt tuning},
  author={Feng, Chun-Mei and Yu, Kai and Liu, Yong and Khan, Salman and Zuo, Wangmeng},
  booktitle={Proceedings of the IEEE/CVF International Conference on Computer Vision},
  pages={2704--2714},
  year={2023}
}

@article{histpt,
  title={Historical test-time prompt tuning for vision foundation models},
  author={Zhang, Jingyi and Huang, Jiaxing and Zhang, Xiaoqin and Shao, Ling and Lu, Shijian},
  journal={Advances in Neural Information Processing Systems},
  volume={37},
  pages={12872--12896},
  year={2024}
}

@article{ctpt,
  title={C-tpt: Calibrated test-time prompt tuning for vision-language models via text feature dispersion},
  author={Yoon, Hee Suk and Yoon, Eunseop and Tee, Joshua Tian Jin and Hasegawa-Johnson, Mark and Li, Yingzhen and Yoo, Chang D},
  journal={arXiv preprint arXiv:2403.14119},
  year={2024}
}

@inproceedings{atpt,
  title={A-TPT: Angular diversity calibration properties for test-time prompt tuning of vision-language models},
  author={Ahamed, Shihab and Miriya Thanthrige, Udaya Sampath K Perera and Rodrigo, Ranga and Khan, Muhammad Haris},
  booktitle={International Conference on Learning Representations},
  volume={2026},
  pages={69175--69199},
  year={2026}
}

@article{promptalign,
  title={Align your prompts: Test-time prompting with distribution alignment for zero-shot generalization},
  author={Abdul Samadh, Jameel and Gani, Mohammad Hanan and Hussein, Noor and Khattak, Muhammad Uzair and Naseer, Muhammad Muzammal and Shahbaz Khan, Fahad and Khan, Salman H},
  journal={Advances in Neural Information Processing Systems},
  volume={36},
  pages={80396--80413},
  year={2023}
}

@inproceedings{tps,
  title={Just shift it: Test-time prototype shifting for zero-shot generalization with vision-language models},
  author={Sui, Elaine and Wang, Xiaohan and Yeung-Levy, Serena},
  booktitle={2025 IEEE/CVF Winter Conference on Applications of Computer Vision (WACV)},
  pages={825--835},
  year={2025},
  organization={IEEE}
}

@article{dpe,
  title={Dual prototype evolving for test-time generalization of vision-language models},
  author={Zhang, Ce and Stepputtis, Simon and Sycara, Katia and Xie, Yaqi},
  journal={Advances in Neural Information Processing Systems},
  volume={37},
  pages={32111--32136},
  year={2024}
}

@article{batclip,
  title={BATCLIP: Bimodal Online Test-Time Adaptation for CLIP},
  author={Maharana, Sarthak Kumar and Zhang, Baoming and Karlinsky, Leonid and Feris, Rogerio and Guo, Yunhui},
  journal={arXiv preprint arXiv:2412.02837},
  year={2024}
}

@inproceedings{panda,
  title={Panda: Test-Time Adaptation with Negative Data Augmentation},
  author={Deng, Ruxi and Bao, Wenxuan and Wei, Tianxin and He, Jingrui},
  booktitle={Proceedings of the AAAI Conference on Artificial Intelligence},
  volume={40},
  number={5},
  pages={3551--3559},
  year={2026}
}

@article{rlcf,
  title={Test-time adaptation with clip reward for zero-shot generalization in vision-language models},
  author={Zhao, Shuai and Wang, Xiaohan and Zhu, Linchao and Yang, Yi},
  journal={arXiv preprint arXiv:2305.18010},
  year={2023}
}

@inproceedings{tda,
  title={Efficient test-time adaptation of vision-language models},
  author={Karmanov, Adilbek and Guan, Dayan and Lu, Shijian and El Saddik, Abdulmotaleb and Xing, Eric},
  booktitle={Proceedings of the IEEE/CVF Conference on Computer Vision and Pattern Recognition},
  pages={14162--14171},
  year={2024}
}

@article{boostadapter,
  title={Boostadapter: Improving vision-language test-time adaptation via regional bootstrapping},
  author={Zhang, Taolin and Wang, Jinpeng and Guo, Hang and Dai, Tao and Chen, Bin and Xia, Shu-Tao},
  journal={Advances in Neural Information Processing Systems},
  volume={37},
  pages={67795--67825},
  year={2024}
}

@inproceedings{dmn,
  title={Dual Memory Networks: A Versatile Adaptation Approach for Vision-Language Models},
  author={Zhang, Yabin and Zhu, Wenjie and Tang, Hui and Ma, Zhiyuan and Zhou, Kaiyang and Zhang, Lei},
  booktitle={Proceedings of the IEEE/CVF conference on computer vision and pattern recognition},
  year={2024}
}

@article{dota,
  title={Dota: Distributional test-time adaptation of vision-language models},
  author={Han, Zongbo and Yang, Jialong and Wang, Guangyu and Li, Junfan and Xu, Qianli and Shou, Mike Zheng and Zhang, Changqing},
  journal={arXiv preprint arXiv:2409.19375},
  year={2024}
}

@inproceedings{oga,
  title={Online Gaussian Test-Time Adaptation of Vision-Language Models.},
  author={Fuchs, Cl{\'e}ment and Zanella, Maxime and De Vleeschouwer, Christophe},
  booktitle={Proceedings of the Computer Vision and Pattern Recognition Conference},
  pages={128--137},
  year={2025}
}

@inproceedings{bca,
  title={Bayesian test-time adaptation for vision-language models},
  author={Zhou, Lihua and Ye, Mao and Li, Shuaifeng and Li, Nianxin and Zhu, Xiatian and Deng, Lei and Liu, Hongbin and Lei, Zhen},
  booktitle={Proceedings of the Computer Vision and Pattern Recognition Conference},
  pages={29999--30009},
  year={2025}
}

@article{zero,
  title={Frustratingly easy test-time adaptation of vision-language models},
  author={Farina, Matteo and Franchi, Gianni and Iacca, Giovanni and Mancini, Massimiliano and Ricci, Elisa},
  journal={Advances in Neural Information Processing Systems},
  volume={37},
  pages={129062--129093},
  year={2024}
}

@inproceedings{calip,
  title={Calip: Zero-shot enhancement of clip with parameter-free attention},
  author={Guo, Ziyu and Zhang, Renrui and Qiu, Longtian and Ma, Xianzheng and Miao, Xupeng and He, Xuming and Cui, Bin},
  booktitle={Proceedings of the AAAI Conference on Artificial Intelligence},
  volume={37},
  number={1},
  pages={746--754},
  year={2023}
}

@inproceedings{mta,
  title={On the test-time zero-shot generalization of vision-language models: Do we really need prompt learning?},
  author={Zanella, Maxime and Ben Ayed, Ismail},
  booktitle={Proceedings of the IEEE/CVF Conference on Computer Vision and Pattern Recognition},
  pages={23783--23793},
  year={2024}
}

@inproceedings{onzeta,
  title={Online zero-shot classification with clip},
  author={Qian, Qi and Hu, Juhua},
  booktitle={European Conference on Computer Vision},
  pages={462--477},
  year={2024},
  organization={Springer}
}

@inproceedings{rtpt,
  title={R-tpt: Improving adversarial robustness of vision-language models through test-time prompt tuning},
  author={Sheng, Lijun and Liang, Jian and Wang, Zilei and He, Ran},
  booktitle={Proceedings of the Computer Vision and Pattern Recognition Conference},
  pages={29958--29967},
  year={2025}
}

@article{ecalp,
  title={Efficient and context-aware label propagation for zero-/few-shot training-free adaptation of vision-language model},
  author={Li, Yushu and Su, Yongyi and Goodge, Adam and Jia, Kui and Xu, Xun},
  journal={arXiv preprint arXiv:2412.18303},
  year={2024}
}

@inproceedings{fan2026moetta,
  title={Moetta: Test-time adaptation under mixed distribution shifts with moe-layernorm},
  author={Fan, Xiao and Jiang, Jingyan and Chen, Zhaoru and Huang, Fanding and Chen, Xiao and Jiang, Qinting and Zhang, Bowen and Tang, Xing and Wang, Zhi},
  booktitle={Proceedings of the AAAI Conference on Artificial Intelligence},
  volume={40},
  number={25},
  pages={21011--21019},
  year={2026}
}

@inproceedings{ttd,
  title={Test-time distillation for continual model adaptation},
  author={Chen, Xiao and Huang, Jiazhen and Liu, Zhiming and Jiang, Qinting and Huang, Fanding and Jiang, Jingyan and Wang, Zhi},
  booktitle={Proceedings of the IEEE/CVF Conference on Computer Vision and Pattern Recognition},
  pages={7593--7604},
  year={2026}
}

@article{TTA_survey_2,
  title={Beyond model adaptation at test time: A survey},
  author={Xiao, Zehao and Snoek, Cees GM},
  journal={arXiv preprint arXiv:2411.03687},
  year={2024}
}

@article{augmix,
  title={Augmix: A simple data processing method to improve robustness and uncertainty},
  author={Hendrycks, Dan and Mu, Norman and Cubuk, Ekin D and Zoph, Barret and Gilmer, Justin and Lakshminarayanan, Balaji},
  journal={arXiv preprint arXiv:1912.02781},
  year={2019}
}

@inproceedings{inet-v2,
  title={Do imagenet classifiers generalize to imagenet?},
  author={Recht, Benjamin and Roelofs, Rebecca and Schmidt, Ludwig and Shankar, Vaishaal},
  booktitle={International conference on machine learning},
  pages={5389--5400},
  year={2019},
  organization={PMLR}
}

@inproceedings{inet-k,
        title={Learning Robust Global Representations by Penalizing Local Predictive Power},
        author={Wang, Haohan and Ge, Songwei and Lipton, Zachary and Xing, Eric P},
        booktitle={Advances in Neural Information Processing Systems},
        pages={10506--10518},
        year={2019}
}

@article{inet-a,
  title={Natural Adversarial Examples},
  author={Dan Hendrycks and Kevin Zhao and Steven Basart and Jacob Steinhardt and Dawn Song},
  journal={CVPR},
  year={2021}
}

@article{inet-r,
  title={The Many Faces of Robustness: A Critical Analysis of Out-of-Distribution Generalization},
  author={Dan Hendrycks and Steven Basart and Norman Mu and Saurav Kadavath and Frank Wang and Evan Dorundo and Rahul Desai and Tyler Zhu and Samyak Parajuli and Mike Guo and Dawn Song and Jacob Steinhardt and Justin Gilmer},
  journal={ICCV},
  year={2021}
}

@INPROCEEDINGS{stanfordcars,
  author={Krause, Jonathan and Stark, Michael and Deng, Jia and Fei-Fei, Li},
  booktitle={2013 IEEE International Conference on Computer Vision Workshops}, 
  title={3D Object Representations for Fine-Grained Categorization}, 
  year={2013},
  }

@inproceedings{food101,
  title     = {Food-101 -- Mining Discriminative Components with Random Forests},
  author    = {Bossard, Lukas and Guillaumin, Matthieu and Van Gool, Luc},
  booktitle = {European Conference on Computer Vision},
  year      = {2014}
}

@techreport{aircraft,
  title={Fine-Grained Visual Classification of Aircraft},
  author={Maji, Subhransu and Rahtu, Esa and Kannala, Juho and Blaschko, Matthew and Vedaldi, Andrea},
  institution={arXiv},
  year={2013}
}

@inproceedings{pets,
  title={Cats and dogs},
  author={Parkhi, Omkar M and Vedaldi, Andrea and Zisserman, Andrew and Jawahar, CV},
  booktitle={2012 IEEE conference on computer vision and pattern recognition},
  year={2012},
}

@inproceedings{flowers102,
  title={Automated flower classification over a large number of classes},
  author={Nilsback, Maria-Elena and Zisserman, Andrew},
  booktitle={2008 Sixth Indian conference on computer vision, graphics \& image processing},
}

@INPROCEEDINGS{sun,
author={J. {Xiao} and J. {Hays} and K. A. {Ehinger} and A. {Oliva} and A. {Torralba} },
booktitle={2010 IEEE Computer Society Conference on Computer Vision and Pattern Recognition},
title={SUN database: Large-scale scene recognition from abbey to zoo},
year={2010},
month={June},}

@inproceedings{dtd,
  title={Describing textures in the wild},
  author={Cimpoi, Mircea and Maji, Subhransu and Kokkinos, Iasonas and Mohamed, Sammy and Vedaldi, Andrea},
  booktitle={Proceedings of the IEEE conference on computer vision and pattern recognition},
  year={2014}
}

@article{eurosat,
  title={Eurosat: A novel dataset and deep learning benchmark for land use and land cover classification},
  author={Helber, Patrick and Bischke, Benjamin and Dengel, Andreas and Borth, Damian},
  journal={IEEE Journal of Selected Topics in Applied Earth Observations and Remote Sensing},
  year={2019},
}

@article{ucf101,
  title={Ucf101: A dataset of 101 human actions classes from videos in the wild},
  author={Soomro, Khurram and Zamir, Amir Roshan and Shah, Mubarak},
  journal={arXiv preprint arXiv:1212.0402},
  year={2012}
}

@article{inet-c,
  title={Benchmarking neural network robustness to common corruptions and perturbations},
  author={Hendrycks, Dan and Dietterich, Thomas},
  journal={arXiv preprint arXiv:1903.12261},
  year={2019}
}

@article{infonce,
  title={Representation learning with contrastive predictive coding},
  author={Oord, Aaron van den and Li, Yazhe and Vinyals, Oriol},
  journal={arXiv preprint arXiv:1807.03748},
  year={2018}
}

@inproceedings{yuan2023robust,
  title={Robust test-time adaptation in dynamic scenarios},
  author={Yuan, Longhui and Xie, Binhui and Li, Shuang},
  booktitle={Proceedings of the IEEE/CVF Conference on Computer Vision and Pattern Recognition},
  pages={15922--15932},
  year={2023}
}

@inproceedings{stata,
  title={Realistic test-time adaptation of vision-language models},
  author={Zanella, Maxime and Fuchs, Cl{\'e}ment and De Vleeschouwer, Christophe and Ben Ayed, Ismail},
  booktitle={Proceedings of the IEEE/CVF Conference on Computer Vision and Pattern Recognition},
  pages={25103--25112},
  year={2025}
}

@inproceedings{vlm_downstream_1,
  title={Open-vocabulary semantic segmentation with mask-adapted clip},
  author={Liang, Feng and Wu, Bichen and Dai, Xiaoliang and Li, Kunpeng and Zhao, Yinan and Zhang, Hang and Zhang, Peizhao and Vajda, Peter and Marculescu, Diana},
  booktitle={Proceedings of the IEEE/CVF conference on computer vision and pattern recognition},
  pages={7061--7070},
  year={2023}
}

@inproceedings{vlm_downstream_2,
  title={Cora: Adapting clip for open-vocabulary detection with region prompting and anchor pre-matching},
  author={Wu, Xiaoshi and Zhu, Feng and Zhao, Rui and Li, Hongsheng},
  booktitle={Proceedings of the IEEE/CVF conference on computer vision and pattern recognition},
  pages={7031--7040},
  year={2023}
}

@article{vlm_survey,
  title={Vision-language models for vision tasks: A survey},
  author={Zhang, Jingyi and Huang, Jiaxing and Jin, Sheng and Lu, Shijian},
  journal={IEEE transactions on pattern analysis and machine intelligence},
  volume={46},
  number={8},
  pages={5625--5644},
  year={2024},
  publisher={IEEE}
}

@inproceedings{imagenet,
  title={Imagenet: A large-scale hierarchical image database},
  author={Deng, Jia and Dong, Wei and Socher, Richard and Li, Li-Jia and Li, Kai and Fei-Fei, Li},
  booktitle={2009 IEEE conference on computer vision and pattern recognition},
  pages={248--255},
  year={2009},
  organization={Ieee}
}

@inproceedings{dirichlet,
  title={Robust test-time adaptation in dynamic scenarios},
  author={Yuan, Longhui and Xie, Binhui and Li, Shuang},
  booktitle={2023 IEEE/CVF Conference on Computer Vision and Pattern Recognition (CVPR)},
  pages={15922--15932},
  year={2023},
  organization={IEEE}
}

@article{moon,
  title={Von Mises-Fisher Mixture Model with Dynamic Shrinkage for Realistic Test-Time Transduction},
  author={Huang, Jiazhen and Liu, Zhiming and Wang, Changhu and Ju, Wei and Qiao, Ziyue and Luo, Xiao},
  journal={arXiv preprint arXiv:2607.15851},
  year={2026}
}

@article{veilleux2021realistic,
  title={Realistic evaluation of transductive few-shot learning},
  author={Veilleux, Olivier and Boudiaf, Malik and Piantanida, Pablo and Ben Ayed, Ismail},
  journal={Advances in Neural Information Processing Systems},
  volume={34},
  pages={9290--9302},
  year={2021}
}

@inproceedings{burns2021limitations,
  title={Limitations of post-hoc feature alignment for robustness},
  author={Burns, Collin and Steinhardt, Jacob},
  booktitle={2021 IEEE/CVF Conference on Computer Vision and Pattern Recognition (CVPR)},
  pages={2525--2533},
  year={2021},
  organization={IEEE}
}

@inproceedings{xiao2025visual,
  title={Visual instance-aware prompt tuning},
  author={Xiao, Xi and Zhang, Yunbei and Li, Xingjian and Wang, Tianyang and Wang, Xiao and Wei, Yuxiang and Hamm, Jihun and Xu, Min},
  booktitle={Proceedings of the 33rd ACM International Conference on Multimedia},
  pages={2880--2889},
  year={2025}
}

@inproceedings{xiao2026layer,
  title={Layer-specific prompt fusion discovery via differentiable search in vision foundation models},
  author={Xiao, Xi and Li, Xingjian and Zhang, Yunbei and Han, Cheng and Liu, Tianming and Wang, Tianyang and Jiang, Runmin and Hamm, Jihun and Wang, Xiao and Xu, Min},
  booktitle={European Conference on Computer Vision},
  pages={391--409},
  year={2026},
  organization={Springer}
}

@article{
xiao2026adapting,
title={Adapting Vision Foundation Models with Cascaded Semantics},
author={Xi Xiao and Xingjian Li and Cheng Han and Tianyang Wang and Lin Zhao and Yunbei Zhang and Guosheng Hu and Runmin Jiang and Xi Li and Xiao Wang and Min Xu},
journal={Transactions on Machine Learning Research},
issn={2835-8856},
year={2026},
url={https://openreview.net/forum?id=SSsobNZJPO},
note={}
}

@article{zhang2026temporal,
  title={Temporal Gains, Spatial Costs: Revisiting Video Fine-Tuning in Multimodal Large Language Models},
  author={Zhang, Linghao and Li, Jungang and Hei, Yonghua and Tao, Sicheng and Dai, Song and Yan, Yibo and Dongfang, Zihao and Liu, Weiting and Qin, Chenxi and Li, Hanqian and others},
  journal={arXiv preprint arXiv:2603.17541},
  year={2026}
}

@article{shi2026androtmem,
  title={AndroTMem: From Interaction Trajectories to Anchored Memory in Long-Horizon GUI Agents},
  author={Shi, Yibo and Li, Jungang and Zhang, Linghao and Dongfang, Zihao and Wu, Biao and Tao, Sicheng and Yan, Yibo and Qin, Chenxi and Liu, Weiting and Lin, Zhixin and others},
  journal={arXiv preprint arXiv:2603.18429},
  year={2026}
}

@inproceedings{chen2025r2ibench,
  title     = {R2I-Bench: Benchmarking Reasoning-Driven Text-to-Image Generation},
  author    = {Chen, Kaijie and Lin, Zihao and Xu, Zhiyang and Shen, Ying and Yao, Yuguang and Rimchala, Joy and Zhang, Jiaxin and Huang, Lifu},
  booktitle = {Proceedings of the 2025 Conference on Empirical Methods in Natural Language Processing},
  pages     = {12595--12630},
  year      = {2025},
  publisher = {Association for Computational Linguistics},
  doi       = {10.18653/v1/2025.emnlp-main.636}
}

@inproceedings{chen2026mvibench,
  title     = {MVI-Bench: A Comprehensive Benchmark for Evaluating Robustness to Misleading Visual Inputs in LVLMs},
  author    = {Chen, Huiyi and Peng, Jiawei and Min, Dehai and Sun, Changchang and Chen, Kaijie and Yan, Yan and Yang, Xu and Cheng, Lu},
  booktitle = {Proceedings of the 43rd International Conference on Machine Learning},
  year      = {2026}
}
}
\nocite{xiao2026layer,xiao2026adapting,shi2026androtmem,zhang2026temporal,chen2025r2ibench,chen2026mvibench}

%%%%%%%%%%%%%%%%%%%%%%%%%%%%%%%%%%%%%%%%%%%%%%%%%%%%%%%%%%%%
\newpage
\appendix
\section{Broader Impacts}
\label{app:impacts}
Our systematic study and the introduction of the TTABC benchmark aim to foster transparency and reliability in the deployment of Vision-Language Models (VLMs) under real-world distribution shifts. 
By demystifying the true drivers of Test-Time Adaptation (TTA), this work might encourage the development of more sustainable, compute-efficient AI systems. 
However, since TTA methods rely heavily on unsupervised signals and streaming test evidence, there is a latent risk that they may amplify pre-existing model biases or suffer from representation collapse when exposed to severely shifted label priors. 
Consequently, we urge practitioners to implement careful monitoring and fairness-aware safeguards when deploying these adaptive systems in high-stakes domains such as healthcare or autonomous navigation.

\section{Limitations and Future Work}
\label{app:limitation}
While our controlled empirical study provides a comprehensive understanding of the current TTA4CLIP landscape, it is inherently bounded by its focus on image classification tasks and the CLIP-based architecture. Real-world applications often involve more complex scenarios, such as dense visual predictions (e.g., object detection, semantic segmentation) and generative multimodal interactions, which remain underexplored in the context of TTA. Furthermore, although we identify that test-time evidence and reliable proxies are the primary drivers of stable adaptation gains, discovering optimal strategies for dynamically selecting the best adaptation paradigm on-the-fly remains an open challenge. Future work should aim to extend the TTABC benchmark to encompass diverse foundational architectures (e.g., autoregressive generative VLMs), broader downstream tasks, and theoretically grounded methods that can automatically adjust their adaptation mechanisms to continuous, unknown distribution shifts.

\section{Large Language Model (LLM) Usage Statement}
\label{app:llm}
We use the LLM as a general-purpose assistant tool. Specifically, the LLM assists in (i) checking grammar and improving clarity of text descriptions, and (ii) suggesting alternative phrasings for some sections. No parts of the paper are generated entirely by the LLM. All research ideas, experiments, model designs, and results are conceived, implemented, and analyzed solely by the authors. The LLM does not contribute to the development of the methodology, experiments, or analysis presented in this paper. We confirm that the use of the LLM is limited to minor writing support and does not constitute a substantive contribution that would qualify it as a co-author.

\section{Detailed Taxonomy of TTA4CLIP Methods}
\label{app:taxonomy}
In Table \ref{tab:taxonomy_details}, we provide a comprehensive taxonomy of all evaluated methods and baselines included in our benchmark. Beyond standard zero-shot and few-shot baselines, we systematically organize TTA4CLIP approaches into our proposed three paradigms: Parameter-based, State-based, and Inference-based methods. 

To clarify the operational assumptions underlying each method, we further characterize them across four key attributes:
\begin{itemize}
    \item \textbf{Zero-Shot}: Whether the method can be deployed directly without requiring any labeled source data for pre-adaptation (e.g., CoOp).
    \item \textbf{Train Req. (Training Required)}: Whether the adaptation process involves gradient-based backpropagation to update model parameters.
    \item \textbf{Sample-wise}: Whether the method adapts independently per single test sample, as opposed to relying on batch-level statistics.
    \item \textbf{Episodic}: Whether the adaptable parameters or external states are strictly reset after each inference step to prevent catastrophic forgetting or error accumulation.
\end{itemize}
This fine-grained breakdown highlights the structural differences, computational requirements, and stability mechanisms of the current TTA4CLIP landscape.

\begin{table}[h]
\centering
\caption{\textbf{Detailed taxonomy of evaluated methods and baselines.} We systematically categorize existing approaches based on their adaptation paradigms. Attributes denote whether a method is zero-shot (requires no source data), requires training (parameter updates), performs sample-wise adaptation, and uses episodic resets. (\ding{51}: Yes, \ding{55}: No, ---: Not Applicable)}
\label{tab:taxonomy_details}
\resizebox{\textwidth}{!}{
\begin{tabular}{llllcccc}
\toprule
\textbf{Paradigm} & \textbf{Category} & \textbf{Method} & \textbf{Venue} & \textbf{Zero-Shot} & \textbf{Train Req.} & \textbf{Sample-wise} & \textbf{Episodic} \\
\midrule
\emph{Zero-Shot} & --- & CLIP & ICML'21 & \ding{51} & --- & --- & --- \\
\midrule
\multirow{6}{*}{\emph{Few-Shot}} & \multirow{3}{*}{Prompt Learning} 
  & CoOp \cite{coop} & IJCV'22 & \ding{55} & \ding{51} & --- & --- \\
& & CoCoOp \cite{cocoop} & CVPR'22 & \ding{55} & \ding{51} & --- & --- \\
& & MaPLe \cite{maple} & CVPR'23 & \ding{55} & \ding{51} & --- & --- \\
\cmidrule(lr){2-8}
& \multirow{3}{*}{TTA Ensemble} 
  & CoOp+TPT & --- & \ding{55} & \ding{51} & \ding{51} & \ding{51} \\
& & CoCoOp+TPT & --- & \ding{55} & \ding{51} & \ding{51} & \ding{51} \\
& & MaPLe+TPT & --- & \ding{55} & \ding{51} & \ding{51} & \ding{51} \\
\midrule
\multirow{11}{*}{\textcolor{param_orange}{\emph{Parameter-based}}} & \multirow{7}{*}{Prompt Learning} 
  & TPT \cite{tpt} & NeurIPS'22 & \ding{51} & \ding{51} & \ding{51} & \ding{51} \\
& & DiffTPT \cite{difftpt} & ICCV'23 & \ding{51} & \ding{51} & \ding{51} & \ding{51} \\
& & HisTPT \cite{histpt} & NeurIPS'24 & \ding{51} & \ding{51} & \ding{51} & \ding{55} \\
& & C-TPT \cite{ctpt} & ICLR'24 & \ding{51} & \ding{51} & \ding{51} & \ding{51} \\
& & A-TPT \cite{atpt} & ICLR'26 & \ding{51} & \ding{51} & \ding{51} & \ding{51} \\
& & R-TPT \cite{rtpt} & CVPR'25 & \ding{51} & \ding{51} & \ding{51} & \ding{51} \\
& & PromptAlign \cite{promptalign} & NeurIPS'23 & \ding{55} & \ding{51} & \ding{51} & \ding{51} \\
& & RLCF \cite{rlcf} & ICLR'24 & \ding{51} & \ding{51} & \ding{51} & \ding{51} \\
\cmidrule(lr){2-8}
& \multirow{2}{*}{Prototype} 
  & TPS \cite{tps} & WACV'25 & \ding{51} & \ding{51} & \ding{51} & \ding{51} \\
& & DPE \cite{dpe} & NeurIPS'24 & \ding{51} & \ding{51} & \ding{51} & \ding{55} \\
% \cmidrule(lr){2-8}
% & Adapter 
\cmidrule(lr){2-8}
& Norm Layer 
  & BATCLIP \cite{batclip} & ICCV'25 & \ding{51} & \ding{51} & \ding{55} & \ding{55} \\
\midrule
\multirow{8}{*}{\textcolor{state_green}{\emph{State-based}}} & \multirow{4}{*}{Cache} 
  & TDA \cite{tda} & CVPR'24 & \ding{51} & \ding{55} & \ding{51} & \ding{55} \\
& & BoostAdapter \cite{boostadapter} & NeurIPS'24 & \ding{51} & \ding{55} & \ding{51} & \ding{55} \\
& & DMN \cite{dmn} & CVPR'24 & \ding{51} & \ding{55} & \ding{51} & \ding{55} \\
& & ECALP \cite{ecalp} & ICLR'25 & \ding{51} & \ding{55} & \ding{51} & \ding{55} \\
\cmidrule(lr){2-8}
& \multirow{4}{*}{Distribution} 
  & DoTA \cite{dota} & NeurIPS'25 & \ding{51} & \ding{55} & \ding{51} & \ding{55} \\
& & OGA \cite{oga} & CVPRW'25 & \ding{51} & \ding{55} & \ding{51} & \ding{55} \\
& & BCA \cite{bca} & CVPR'25 & \ding{51} & \ding{55} & \ding{51} & \ding{55} \\
& & OnZeta \cite{onzeta} & ECCV'24 & \ding{51} & \ding{55} & \ding{51} & \ding{55} \\
\midrule
\multirow{4}{*}{\textcolor{infer_blue}{\emph{Inference-based}}} & \multirow{2}{*}{View Aggregation} 
  & ZERO \cite{zero} & NeurIPS'24 & \ding{51} & \ding{55} & \ding{51} & \ding{51} \\
& & MTA \cite{mta} & CVPR'24 & \ding{51} & \ding{55} & \ding{51} & \ding{51} \\
\cmidrule(lr){2-8}
& \multirow{2}{*}{Feature Modulation} 
  & CALIP \cite{calip} & AAAI'23 & \ding{51} & \ding{55} & \ding{51} & \ding{51} \\
& & Panda \cite{panda} & AAAI'26 & \ding{51} & \ding{55} & \ding{55} & \ding{55} \\
\bottomrule
\end{tabular}
}
\end{table}

\section{Experimental Details of Datasets, Baselines, and Hyperparameters}
\label{app:detail}

\subsection{Datasets}
\label{app:base-data}

To comprehensively evaluate the generalization capability of VLMs under various distribution shifts, we utilize 15 diverse image classification datasets, categorized into four main groups:

\noindent\textbf{1. Original ImageNet (In-Distribution):}
\begin{itemize}[leftmargin=*, parsep=0pt, itemsep=2pt, topsep=2pt]
    \item \textbf{ImageNet} \cite{imagenet}: ImageNet is used as the in-distribution evaluation benchmark. We evaluate models on the standard ImageNet-1K validation set, which contains 50,000 images from 1,000 object categories. This dataset serves as the reference test distribution for assessing standard image classification performance under the original ImageNet data distribution.
\end{itemize}

\noindent\textbf{2. Natural Distribution Shifts (Out-of-Distribution):}
\begin{itemize}[leftmargin=*, parsep=0pt, itemsep=2pt, topsep=2pt]
    \item \textbf{ImageNet-V2} \cite{inet-v2}: ImageNet-V2 is used as a natural distribution-shift test set for ImageNet classification. It contains newly collected images from the same 1,000 ImageNet classes, with 10,000 images in each test set variant. Since the images are collected independently from the original ImageNet validation set while following a similar collection protocol, this benchmark evaluates robustness to natural changes in data sampling and collection procedures.
    \item \textbf{ImageNet-Sketch} \cite{inet-k}: ImageNet-Sketch is used to evaluate robustness to a strong modality and style shift from natural photographs to sketch images. It consists of sketch-style images covering the 1,000 ImageNet classes, with approximately 50 images per class. Because sketches remove most color and texture cues while preserving object shape, this dataset tests whether models can generalize beyond photo-realistic visual appearances.
    \item \textbf{ImageNet-A} \cite{inet-a}: ImageNet-A is used as a challenging natural adversarial test set. It contains natural, unmodified real-world images from 200 ImageNet classes that are difficult for standard ImageNet-trained models to classify correctly. This benchmark evaluates model robustness under naturally hard examples involving unusual object appearances, viewpoints, backgrounds, or visual contexts.
    \item \textbf{ImageNet-R} \cite{inet-r}: ImageNet-R is used to test robustness to non-photorealistic visual renditions. It contains 30,000 images from 200 ImageNet classes, including artistic forms such as cartoons, paintings, sculptures, graphics, embroidery, origami, toys, and other renditions. This dataset evaluates whether models can preserve category recognition when object appearance changes substantially in style and texture.
\end{itemize}

\begin{table*}[h]
\centering
\small
\caption{\textbf{Detailed Results on ImageNet-C (ViT-B/16)\cite{vit}.} Accuracy (\%) across 15 corruption types at severity level 5. The types are grouped into four main categories: Noise, Blur, Weather, and Digital.}
\label{tab:imagenetc_vit}
\setlength{\tabcolsep}{3pt} % 稍微收紧列间距
\begin{adjustbox}{width=\textwidth}
\begin{tabular}{l | cccc | ccc | cccc | cccc | c}
\toprule
\multirow{2}{*}{\textbf{Method}} & \multicolumn{4}{c|}{\textbf{Noise}} & \multicolumn{3}{c|}{\textbf{Blur}} & \multicolumn{4}{c|}{\textbf{Weather}} & \multicolumn{4}{c|}{\textbf{Digital}} & \multirow{2}{*}{\textbf{Avg.}} \\
\cmidrule(lr){2-5} \cmidrule(lr){6-8} \cmidrule(lr){9-12} \cmidrule(lr){13-16}
& \textbf{Gauss.} & \textbf{Shot} & \textbf{Impul.} & \textbf{Defoc.} & \textbf{Glass} & \textbf{Motion} & \textbf{Zoom} & \textbf{Snow} & \textbf{Frost} & \textbf{Fog} & \textbf{Brit.} & \textbf{Contr.} & \textbf{Elastic} & \textbf{Pixel} & \textbf{JPEG} & \\
\midrule
CLIP & 13.25 & 14.16 & 13.49 & 24.23 & 15.72 & 24.47 & 22.63 & 33.09 & 31.09 & 37.60 & 55.61 & 17.09 & 13.43 & 33.04 & 33.70 & 25.51 \\
\midrule
CoOp & 16.29 & 17.27 & 15.85 & 26.91 & 17.89 & 27.82 & 25.61 & 35.95 & 34.33 & 40.08 & 60.58 & 20.52 & 15.11 & 38.24 & 37.51 & 28.66 \\
CoOp+TPT & 15.72 & 16.67 & 15.36 & 27.28 & 18.24 & 28.14 & 26.42 & 36.80 & 35.17 & 41.06 & 61.28 & 22.21 & 16.03 & 39.92 & 38.51 & 29.25 \\
CoCoOp & 14.48 & 15.41 & 14.08 & 21.58 & 13.19 & 25.00 & 23.49 & 28.03 & 27.97 & 35.40 & 53.39 & 15.87 & 13.29 & 29.15 & 30.85 & 24.08 \\
CoCoOp+TPT & 14.63 & 15.66 & 14.28 & 22.23 & 13.66 & 25.20 & 23.82 & 28.80 & 28.72 & 35.92 & 54.20 & 16.45 & 13.55 & 30.35 & 31.83 & 24.62 \\
MaPLe & 15.58 & 16.71 & 15.83 & 26.19 & 17.14 & 26.97 & 24.99 & 33.29 & 33.34 & 38.56 & 57.82 & 20.10 & 14.79 & 36.76 & 35.87 & 27.60 \\
MaPLe+TPT & 15.45 & 16.65 & 15.81 & 26.39 & 17.25 & 27.14 & 25.18 & 33.59 & 33.64 & 38.79 & 58.09 & 20.48 & 14.99 & 37.23 & 36.15 & 27.79 \\
\midrule
TPT & 8.86 & 9.63 & 9.67 & 24.50 & 16.14 & 24.85 & 23.52 & 34.47 & 32.57 & 38.74 & 56.30 & 19.40 & 14.25 & 35.35 & 35.03 & 25.55 \\
C-TPT & 12.95 & 13.57 & 13.48 & 23.19 & 14.93 & 24.21 & 22.42 & 32.96 & 30.19 & 36.63 & 53.69 & 17.31 & 12.65 & 31.96 & 31.88 & 24.80 \\
A-TPT & 8.65 & 9.40 & 9.54 & 24.29 & 15.93 & 24.65 & 23.52 & 33.95 & 32.24 & 38.13 & 55.84 & 18.75 & 14.41 & 34.31 & 34.26 & 25.19 \\
DiffTPT & 10.17 & 11.57 & 10.33 & 24.03 & 17.27 & 25.13 & 24.13 & 35.37 & 32.13 & 38.90 & 57.07 & 18.37 & 16.43 & 34.80 & 35.30 & 26.07 \\
HisTPT & 13.15 & 14.08 & 13.27 & 24.14 & 15.58 & 24.17 & 22.32 & 32.80 & 31.03 & 37.54 & 55.38 & 17.19 & 13.37 & 33.16 & 33.47 & 25.38 \\
PromptAlign & 13.91 & 15.31 & 14.75 & 27.28 & 18.31 & 27.97 & 26.85 & 34.80 & 34.86 & 39.53 & 58.70 & 22.08 & 18.21 & 40.42 & 38.33 & 28.75 \\
RLCF & 13.29 & 14.24 & 13.51 & 24.55 & 15.99 & 24.77 & 22.93 & 33.65 & 31.62 & 38.16 & 56.32 & 17.47 & 13.77 & 33.84 & 34.28 & 25.89 \\
TPS & 11.04 & 12.27 & 10.65 & 24.42 & 16.11 & 24.82 & 25.41 & 35.30 & 34.05 & 39.61 & 55.99 & 29.84 & 17.05 & 38.70 & 36.18 & 27.43 \\
DPE & 6.82 & 8.09 & 6.33 & 27.44 & 19.00 & 27.97 & 27.18 & 37.58 & 35.59 & 42.20 & 58.96 & 22.76 & 18.74 & 39.67 & 37.75 & 27.74 \\
BATCLIP & 19.37 & 21.85 & 20.11 & 24.23 & 21.87 & 29.60 & 29.26 & 36.09 & 33.67 & 42.95 & 56.47 & 25.62 & 27.46 & 37.76 & 37.41 & 30.91 \\
\midrule
TDA & 15.73 & 16.64 & 16.32 & 26.03 & 17.60 & 26.60 & 24.65 & 35.61 & 33.55 & 40.07 & 57.71 & 19.29 & 15.93 & 36.19 & 35.94 & 27.86 \\
BoostAdapter & 15.77 & 16.74 & 16.41 & 26.10 & 17.73 & 26.80 & 24.81 & 35.82 & 33.94 & 40.30 & 57.93 & 19.37 & 16.19 & 36.43 & 36.04 & 28.03 \\
% Tent & 1.99 & 2.20 & 2.90 & 27.78 & 23.21 & 30.38 & 25.76 & 36.75 & 32.29 & 41.97 & 56.78 & 27.79 & 13.77 & 40.37 & 40.20 & 26.94 \\
% SAR & 22.44 & 23.56 & 22.92 & 27.14 & 24.33 & 30.36 & 26.51 & 36.50 & 34.32 & 42.11 & 56.54 & 27.78 & 20.37 & 39.68 & 39.78 & 31.62 \\
% DeYO & 24.45 & 26.21 & 26.30 & 28.72 & 27.35 & 33.80 & 30.87 & 39.29 & 36.06 & 45.17 & 57.50 & 33.24 & 29.47 & 43.20 & 43.00 & 34.98 \\
DMN & 13.15 & 14.05 & 13.40 & 24.44 & 15.94 & 24.65 & 22.87 & 33.40 & 31.45 & 38.16 & 56.26 & 17.54 & 13.81 & 33.63 & 34.22 & 25.80 \\
DoTA & 5.70 & 6.52 & 5.32 & 22.91 & 15.61 & 22.81 & 24.32 & 34.67 & 33.73 & 39.54 & 55.71 & 28.90 & 18.06 & 38.54 & 34.87 & 25.81 \\
OGA & 15.17 & 16.06 & 15.72 & 25.25 & 17.18 & 26.15 & 24.31 & 34.88 & 33.02 & 39.74 & 57.64 & 19.09 & 15.96 & 35.28 & 35.14 & 27.37 \\
BCA & 0.12 & 0.14 & 0.12 & 2.39 & 1.13 & 1.96 & 4.20 & 2.83 & 4.41 & 10.03 & 23.02 & 1.22 & 1.75 & 2.56 & 3.11 & 3.93 \\
OnZeta & 16.59 & 17.61 & 17.36 & 27.83 & 18.87 & 28.70 & 26.74 & 37.30 & 35.21 & 42.15 & 60.73 & 18.97 & 17.05 & 37.77 & 37.86 & 29.38 \\
ECALP & 17.45 & 18.57 & 18.00 & 27.76 & 19.00 & 28.43 & 26.54 & 37.81 & 35.60 & 42.38 & 60.14 & 21.20 & 17.38 & 38.47 & 37.74 & 29.76 \\
\midrule
ZERO & 3.78 & 4.49 & 3.73 & 20.32 & 13.19 & 19.54 & 22.19 & 31.48 & 30.81 & 36.58 & 52.89 & 23.46 & 15.01 & 35.26 & 31.51 & 22.95 \\
MTA & 9.20 & 9.37 & 8.82 & 23.92 & 15.53 & 23.94 & 23.11 & 34.01 & 32.70 & 38.70 & 56.15 & 20.74 & 14.63 & 35.38 & 35.08 & 25.42 \\
CALIP & 14.84 & 15.80 & 14.93 & 25.51 & 17.16 & 25.96 & 24.00 & 34.30 & 33.13 & 38.92 & 57.57 & 18.67 & 14.83 & 35.56 & 35.75 & 27.13 \\
Panda & 15.33 & 16.36 & 15.56 & 25.62 & 16.78 & 26.25 & 24.04 & 33.90 & 32.19 & 38.90 & 55.53 & 18.66 & 14.85 & 34.99 & 34.40 & 26.89 \\
\bottomrule
\end{tabular}
\end{adjustbox}
\end{table*}

\noindent\textbf{3. Fine-grained Classification Shifts:}
\begin{itemize}[leftmargin=*, parsep=0pt, itemsep=2pt, topsep=2pt]
    \item \textbf{StanfordCars} \cite{stanfordcars}: StanfordCars is used as a fine-grained evaluation benchmark for automobile recognition. The dataset contains 16,185 images from 196 car categories, with an official test split of 8,041 images. Each category corresponds to a specific car make, model, and year, requiring models to distinguish subtle differences in vehicle shape, design, and local visual details.
    \item \textbf{Food101} \cite{food101}: Food101 is used as a fine-grained food classification benchmark. It contains 101 food categories with 1,000 images per category, including 250 manually reviewed test images for each class. The dataset is challenging because food images often exhibit large intra-class variation due to differences in ingredients, preparation styles, lighting, viewpoints, and presentation.
    \item \textbf{FGVC Aircraft} \cite{aircraft}: FGVC Aircraft is used as a fine-grained aircraft recognition benchmark. The commonly used benchmark contains 10,000 images from 100 aircraft variants, with images organized by visually similar aircraft models. This dataset requires models to identify subtle structural differences, such as wing shape, engine placement, fuselage design, and other fine-grained aircraft-specific cues.
    \item \textbf{OxfordPets} \cite{pets}: OxfordPets is used as a fine-grained pet breed classification benchmark. It contains 37 pet categories, with roughly 200 images per category, covering different breeds of cats and dogs. The images vary significantly in scale, pose, and lighting, making the dataset useful for evaluating fine-grained animal recognition under diverse visual conditions.
    \item \textbf{Flowers102} \cite{flowers102}: Flowers102 is used as a fine-grained flower classification benchmark. It contains 102 flower categories commonly found in the United Kingdom, with each category containing between 40 and 258 images. The dataset requires models to distinguish visually similar flower species based on subtle differences in color, shape, petal structure, and local appearance.
    \item \textbf{SUN397} \cite{sun}: SUN397 is used as a scene recognition evaluation benchmark. It contains 397 scene categories and 108,754 images, covering diverse indoor, outdoor, natural, and man-made environments. Unlike object-centric datasets, SUN397 evaluates the ability of models to recognize scene-level concepts based on global layout, contextual cues, and object co-occurrence.
    \item \textbf{DTD} \cite{dtd}: DTD is used as a texture recognition benchmark. It contains 5,640 images from 47 describable texture categories, with 120 images per category. Instead of focusing on object identity, this dataset evaluates whether models can recognize visual attributes such as striped, dotted, cracked, woven, porous, or other texture patterns.
    \item \textbf{EuroSAT} \cite{eurosat}: EuroSAT is used as a remote sensing classification benchmark. It contains 27,000 labeled satellite images from 10 land use and land cover classes, constructed from Sentinel-2 satellite imagery. This dataset introduces a domain shift from natural ground-level images to overhead remote sensing images, requiring models to classify land patterns such as residential areas, forests, rivers, highways, and agricultural regions.
    \item \textbf{UCF101} \cite{ucf101}: UCF101 is used as an action recognition benchmark adapted to our image-based evaluation setting. The original dataset consists of realistic videos from 101 human action categories collected from YouTube. In our protocol, we use the middle frame of each video as the image input, so the task evaluates whether models can infer action-related semantics from static visual cues such as human pose, objects, and scene context.
\end{itemize}

\noindent\textbf{4. Corruption Shifts:}
\begin{itemize}[leftmargin=*, parsep=0pt, itemsep=2pt, topsep=2pt]
    \item \textbf{ImageNet-C} \cite{inet-c}: ImageNet-C is used to evaluate robustness to common image corruptions. It is constructed by applying 15 types of algorithmic corruptions to ImageNet validation images, with each corruption applied at 5 severity levels. The corruptions include noise, blur, weather effects, and digital distortions, making this benchmark suitable for assessing model reliability under degraded visual conditions that may occur in real-world deployment.
\end{itemize}

\noindent\textbf{5. Label Shifts:}
\begin{itemize}[leftmargin=*, parsep=0pt, itemsep=2pt, topsep=2pt]
    \item \textbf{Temporal Correlation}: Following prior TTA works \cite{yuan2023robust}, we simulate semantic label shifts by sampling using a Dirichlet distribution $\text{Dir}(\alpha)$, where a smaller $\alpha$ indicates a more severe class imbalance. Details are provided in App.\ref{app:label-shift}.
\end{itemize}

\begin{table*}[h]
\centering
\small
\caption{\textbf{Main Results on Natural and Corruption Shifts.} Accuracy (\%) is reported for CLIP ResNet-50 \cite{resnet}. }
\label{tab:rn50-natural}
\begin{adjustbox}{width=0.8\textwidth}
\begin{tabular}{l c | cccc c | c}
\toprule
\multirow{2}{*}{\textbf{Method}} & \textbf{Original} & \multicolumn{5}{c|}{\textbf{Natural Shifts}} & \textbf{Corruption} \\
\cmidrule(lr){2-2} \cmidrule(lr){3-7} \cmidrule(lr){8-8}
& \textbf{INet} & \textbf{I-V2} & \textbf{I-A} & \textbf{I-R} & \textbf{I-S} & \textbf{Avg.} & \textbf{Avg.} \\
\midrule
CLIP & 58.15 & 51.51 & 21.84 & 56.12 & 33.35 & 40.71 & 11.26 \\
\midrule
CoOp & 63.47 & 55.66 & 23.19 & 56.72 & 34.58 & 42.54 & 12.70 \\
CoCoOp & 61.74 & 54.58 & 24.33 & 57.09 & 34.39 & 42.60 & 12.29 \\
TPT & 60.24 & 53.86 & 23.92 & 58.24 & 34.93 & 42.74 & 11.66 \\
CoOp+TPT & 64.68 & 57.14 & 24.60 & 58.00 & 35.51 & 43.81 & 12.99 \\
CoCoOp+TPT & 62.45 & 55.33 & 25.04 & 57.86 & 34.92 & 43.29 & 12.44 \\
DiffTPT & 59.87 & 54.90 & 24.13 & 56.40 & 35.18 & 42.65 & 12.03 \\
HisTPT & 57.73 & 51.51 & 21.88 & 56.25 & 33.38 & 40.76 & 11.14 \\
C-TPT & 60.00 & 53.81 & 22.77 & 57.47 & 34.17 & 42.06 & 12.25 \\
A-TPT & 60.74 & 54.68 & 24.91 & 57.47 & 35.27 & 43.08 & - \\
TPS & 61.56 & 54.94 & 29.20 & 62.27 & 37.20 & 45.90 & 12.81 \\
DPE & 61.76 & 54.58 & 27.64 & 59.16 & 38.18 & 44.89 & - \\
BATCLIP & 36.36 & 37.60 & 9.12 & 27.59 & 5.20 & 19.88 & 3.20 \\
RLCF & 59.12 & 52.33 & 22.51 & 57.23 & 34.08 & 41.54 & 11.53 \\
TDA & 59.78 & 52.10 & 22.56 & 57.19 & 35.76 & 41.90 & 12.47 \\
BoostAdapter & 59.85 & 51.98 & 22.56 & 57.69 & 36.08 & 42.08 & 12.57 \\
DMN & 58.94 & 51.52 & 22.09 & 56.44 & 34.44 & 41.12 & 11.47 \\
DoTA & 61.41 & 53.59 & 27.95 & 58.16 & 36.30 & 44.00 & 12.42 \\
OGA & 59.30 & 49.51 & 22.52 & 57.13 & 35.14 & 41.08 & 12.21 \\
BCA & 58.97 & 53.37 & 26.24 & 39.54 & 10.58 & 32.43 & 2.92 \\
ZERO & 59.86 & 54.07 & 26.15 & 57.36 & 33.87 & 42.86 & 10.40 \\
OnZeta & 62.69 & 54.04 & 23.11 & 60.54 & 38.75 & 44.11 & 13.65 \\
ECALP & 61.25 & 53.30 & 23.32 & 60.49 & 38.69 & 43.95 & 14.02 \\
MTA & 60.21 & 53.80 & 25.72 & 58.50 & 34.94 & 43.24 & 11.20 \\
\bottomrule
\end{tabular}
\end{adjustbox}
\end{table*}

\subsection{Construction of Temporally Correlated Label-Shift Streams}
\label{app:label-shift}
We generate non-i.i.d. test streams using a Dirichlet-based protocol to evaluate the robustness of online TTA methods under temporal label correlation~\cite{yuan2023robust}. 
Given a test set with \(K\) classes, we first divide the stream into multiple temporal slots. 
For each class, its samples are allocated across slots according to a Dirichlet distribution,
\[
\mathbf{p}_k \sim \mathrm{Dir}(\alpha \mathbf{1}),
\]
where \(\mathbf{p}_k\) denotes the slot allocation probability for class \(k\). 
The concentration parameter \(\alpha\) controls the strength of temporal correlation. 
A larger \(\alpha\) produces a nearly i.i.d. stream, while a smaller \(\alpha\) concentrates samples of the same class into fewer slots, leading to stronger temporal dependency.

In addition, we consider a separate-class stream as the most extreme case, corresponding to the limit \(\alpha \to 0\). 
In this setting, classes are randomly permuted, and all samples from the same class appear contiguously before the stream transitions to the next class. 
This protocol allows us to isolate the effect of temporal label correlation without changing the image distribution or the label set.

\subsection{Implementation Details and Hyperparameters}
\label{app:method-details}

To ensure a fair and controlled comparison, we maintain consistent default hyperparameters across all applicable methods, while following their original implementations for specific configurations. Unless otherwise specified, we use the CLIP ViT-B/16 as the default backbone. All input images are resized to $224 \times 224$ and normalized according to the pre-trained CLIP statistics. 

For text representations, class prototypes are initialized with the standard template \texttt{``a photo of a [CLS]''}, where \texttt{[CLS]} denotes the corresponding class name. Specifically for prompt-tuning methods, we initialize 4 learnable context tokens with \texttt{``a\_photo\_of\_a''}. 

During TTA, parameter-based methods use the SGD optimizer by default, with a learning rate of $1\times10^{-4}$ for norm-layer updates and $5\times10^{-3}$ for others, and we perform an episodic reset per sample for most methods. For methods requiring data augmentation, we set the number of augmented views to $N=64$ and filter the top 10\% ($\rho=0.1$) of views based on prediction confidence. 

For batch-wise methods and online adaptation protocols, we adopt a default batch size of $64$ and process the test stream in a strictly causal manner. Finally, all experiments are conducted with the same random seed to ensure reproducibility, and are evaluated on Tesla V100S-PCIE-32GB GPUs to accurately measure inference latency and peak memory consumption.

\subsubsection{Parameter-Based Methods}

\noindent\textbf{Prompt Learning:}
\begin{itemize}[leftmargin=*, parsep=0pt, itemsep=2pt, topsep=2pt]
    \item \textbf{TPT} \cite{tpt}: For each test image, AugMix produces 63 augmented views, forming a batch of 64 together with the original image. The prompt is updated by marginal entropy minimization on the top 10\% most confident views.
    \item \textbf{DiffTPT} \cite{difftpt}: We follow the diffusion-augmented prompt tuning setting with 32 diffusion-generated views. The diffusion guidance scale is set to $3.0$ and the number of diffusion steps is $10$. Candidate views are filtered using a cosine-similarity selection ratio of $0.8$ and a self-entropy selection ratio of $0.3$.
    \item \textbf{HisTPT} \cite{histpt}: We use the original sequential adaptation setting without episodic reset. The local knowledge bank size is set to $32$, the number of hard-sample features is $16$, and the global knowledge bank is updated with EMA momentum $0.99$. HisTPT uses a batch size of $1$ while retaining the same 4-token prompt initialization.
    \item \textbf{C-TPT} \cite{ctpt}: C-TPT inherits the same adaptation pipeline as TPT, including confidence-based view selection, marginal-entropy minimization, batch size 64, and one-step prompt update with SGD. In addition, it introduces a text feature dispersion regularizer weighted by $\lambda$, which encourages larger dispersion among class text features. We set the regularization weight to $\lambda=50.0$, and disable the optional two-step optimization scheme.
    \item \textbf{A-TPT} \cite{atpt}: A-TPT also builds on the TPT framework and preserves the same confidence-based view filtering and entropy-driven adaptation procedure. Its key additional parameter is the angular diversity regularization weight $\lambda$, together with a cosine-similarity clamp threshold $\tau$ used in the angular constraint. We set $\tau=0.9999$ and optimize with AdamW. Following the original paper, we use $\lambda=10.0$ on natural shift datasets and $\lambda=80.0$ on fine-grained datasets.
    \item \textbf{R-TPT} \cite{rtpt}: R-TPT adopts the standard prompt adaptation stage with top-10\% confident view selection, but replaces the entropy objective with average entropy minimization and further incorporates adversarially perturbed inputs during evaluation. We use SGD with learning rate $5\times10^{-3}$ and one adaptation step. For ViT backbones, the PGD perturbation budget is set to $\epsilon=4/255$ with one attack step, and the attack step size is $\epsilon/4$. After adaptation, predictions from multiple views are fused using similarity-based weighting.
    \item \textbf{RLCF} \cite{rlcf}: RLCF follows the TPT-style adaptation pipeline but replaces the entropy objective with reward-weighted optimization guided by an auxiliary CLIP reward model. We sample the top $K=3$ candidate classes for each selected view, and compute rewards using a ViT-L/14 reward architecture. Reward centering is enabled, reward standardization is disabled, and reward processing is performed per sample rather than across the batch. We use the weighted reward-score setting, while keeping entropy regularization and multi-reward-model ensembling disabled.

\end{itemize}

\noindent\textbf{Prototype Modulation:}
\begin{itemize}[leftmargin=*, parsep=0pt, itemsep=2pt, topsep=2pt]
    \item \textbf{TPS} \cite{tps}: We use prototype prompts instead of learnable textual prompts. Textual prototypes are constructed per label with the \texttt{gpt4\_x\_templates} prototype type, combining template-based descriptions with GPT-4 generated concepts. During adaptation, 64-view batches are used and the top 10\% most confident views are selected.
    \item \textbf{DPE} \cite{dpe}: We use a positive cache with shot capacity $3$, cache weight $\alpha=6.0$, and similarity temperature $\beta=5.0$. Text and image prototype residuals are optimized with learning rates $5\times10^{-3}$, and the InfoNCE alignment loss is weighted by $0.5$. The global prototype update threshold is set to $0.1$.
\end{itemize}

\noindent\textbf{Adapter \& Norm Layer Tuning:}
\begin{itemize}[leftmargin=*, parsep=0pt, itemsep=2pt, topsep=2pt]
    \item \textbf{PromptAlign} \cite{promptalign}: We adapt MaPLe prompts using $n\_ctx=2$ and prompt depth $3$, initialized from the released MaPLe checkpoint. AdamW is used for optimization. Both TPT loss and distribution alignment are enabled, with confidence thresholds $0.1$ for TPT and alignment. The distribution alignment loss weight is $100.0$, and alignment is applied from layer $0$ to layer $3$ using pre-computed ImageNet visual statistics.
    \item \textbf{BATCLIP} \cite{batclip}: We tune normalization parameters in a continuous, non-episodic manner. AdamW is used with learning rate $1\times10^{-4}$ and batch size $64$.
    \item \textbf{Tent} \cite{Tent}: We optimize the affine parameters of normalization layers by entropy minimization. The learning rate is $1\times10^{-4}$, the batch size is $64$, and episodic reset is disabled.
    \item \textbf{SAR} \cite{SAR}: We use continuous normalization-layer adaptation with learning rate $1\times10^{-4}$. The entropy filtering margin multiplier is $e_0=0.40$, scaled by $\log C$, and the model recovery threshold is $e_m=0.2$.
    \item \textbf{DeYO} \cite{DEYO}: We use learning rate $1\times10^{-4}$ and continuous adaptation. The DeYO entropy margin multiplier is $0.50$, and the base margin is $0.40$, both scaled by $\log C$. For PLPD, we use patch-based augmentation with patch length $4$, occlusion size $112$, and starting coordinates $(56,56)$. The PLPD threshold is $0.3$, with both entropy and PLPD reweighting enabled.
\end{itemize}

\subsubsection{State-Based Methods}

\noindent\textbf{Cache:}
\begin{itemize}[leftmargin=*, parsep=0pt, itemsep=2pt, topsep=2pt]
     \item \textbf{TDA} \cite{tda}: We use both positive and negative caches. The positive cache has shot capacity $3$, $\alpha=2.0$, and $\beta=5.0$. The negative cache has shot capacity $2$, $\alpha=0.117$, and $\beta=1.0$, with entropy thresholds in $[0.2,0.5]$ and mask thresholds in $[0.03,1.0]$. Augmented views are filtered using the top 10\% confidence criterion.
    \item \textbf{BoostAdapter} \cite{boostadapter}: We retain the same positive and negative cache configuration as TDA. In addition, inference on the original image is enabled and the delta term is set to $0$, following the default regional bootstrapping configuration.
    \item \textbf{DMN} \cite{dmn}: We set the memory size to $50$ and the memory sharpness parameter to $\beta=5.5$. The text logit weight is $\beta_2=1.0$ and the memory logit weight is $\beta_3=0.1$. The mapping module uses the \texttt{bias} form and is applied to all positions without shared parameters.
    \item \textbf{ECALP} \cite{ecalp}: We run ECALP in a streaming, non-episodic setting with batch size $1$ and prompt templates enabled. The label-propagation graph uses $k_{\text{text}}=3$ text neighbors and $k_{\text{image}}=8$ image neighbors. We set $\gamma=10.0$, $\alpha=1.0$, $\beta=0.2$, and perform $3$ propagation iterations.

\end{itemize}

\begin{table*}[h]
\centering
\scriptsize
\caption{\textbf{Main Results on Fine-grained Shifts.} Accuracy (\%) reported for CLIP ResNet-50.}
\label{tab:rn50-fine}
\begin{adjustbox}{width=\textwidth}
\begin{tabular}{l c | cccccccccc | c}
\toprule
\multirow{2}{*}{\textbf{Method}} & \textbf{Original} & \multicolumn{10}{c|}{\textbf{Fine-grained Datasets}} & \textbf{Overall} \\
\cmidrule(lr){2-2} \cmidrule(lr){3-12} \cmidrule(lr){13-13}
& \textbf{INet} & \textbf{Cal} & \textbf{Pets} & \textbf{Cars} & \textbf{FLW} & \textbf{Food} & \textbf{Air} & \textbf{SUN} & \textbf{DTD} & \textbf{SAT} & \textbf{UCF} & \textbf{Avg.} \\
\midrule
Zero-shot CLIP & 58.15 & 85.72 & 83.62 & 55.73 & 61.67 & 75.37 & 15.63 & 58.80 & 40.37 & 23.67 & 58.82 & 55.94 \\
\midrule
CoOp & 63.47 & 86.65 & 86.97 & 55.42 & 61.71 & 77.13 & 15.06 & 58.15 & 37.29 & 26.36 & 59.03 & 56.38 \\
CoCoOp & 61.74 & 88.40 & 87.30 & 54.68 & 65.21 & 77.42 & 15.42 & 59.40 & 38.71 & 27.32 & 59.79 & 57.37 \\
TPT & 60.24 & 87.14 & 84.68 & 57.69 & 62.44 & 76.48 & 17.31 & 61.26 & 41.02 & 24.21 & 60.22 & 57.25 \\
CoOp+TPT & 64.68 & 87.06 & 87.38 & 56.34 & 61.31 & 77.91 & 15.90 & 59.65 & 38.65 & 26.54 & 60.61 & 57.14 \\
CoCoOp+TPT & 62.45 & 88.48 & 87.93 & 55.59 & 65.53 & 77.98 & 15.39 & 60.32 & 39.48 & 27.73 & 60.61 & 57.90 \\
DiffTPT & 59.87 & 87.26 & 85.31 & 57.68 & 62.81 & 76.90 & 16.47 & 61.71 & 41.67 & 27.19 & 61.09 & 57.81 \\
HisTPT & 57.73 & 85.68 & 83.48 & 55.88 & 61.59 & 75.51 & 15.60 & 58.66 & 40.07 & 23.58 & 58.84 & 55.89 \\
C-TPT & 60.00 & 86.69 & 83.84 & 56.59 & 65.25 & 76.20 & 17.91 & 60.46 & 41.61 & 26.09 & 58.90 & 57.35 \\
A-TPT & 60.74 & 84.22 & 83.59 & 54.41 & 61.51 & 74.02 & 16.02 & 59.69 & 40.60 & 22.17 & 58.52 & 55.48 \\
TPS & 61.56 & 88.40 & 78.41 & 58.29 & 66.46 & 73.07 & 19.08 & 58.82 & 46.39 & 24.86 & 61.27 & 57.51 \\
DPE & 61.76 & 88.36 & 85.39 & 59.22 & 63.99 & 77.22 & 16.89 & 63.06 & 42.49 & 23.69 & 63.26 & 58.36 \\
BATCLIP & 36.36 & 62.52 & 45.27 & 6.22 & 8.00 & 5.56 & 3.69 & 35.79 & 20.45 & 14.15 & 36.80 & 23.85 \\
Panda & 45.16 & 63.37 & 48.84 & 8.77 & 11.04 & 14.07 & 4.35 & 40.66 & 21.22 & 14.15 & 37.80 & 26.43 \\
RLCF & 59.12 & 86.45 & 84.38 & 56.45 & 62.40 & 76.10 & 16.17 & 59.57 & 41.19 & 23.93 & 60.51 & 56.72 \\
TDA & 59.78 & 88.07 & 83.95 & 56.88 & 65.61 & 76.28 & 16.50 & 60.57 & 41.49 & 31.74 & 61.30 & 58.24 \\
BoostAdapter & 59.85 & 87.91 & 83.89 & 57.17 & 65.61 & 76.47 & 15.90 & 60.99 & 41.78 & 32.33 & 61.62 & 58.37 \\
DMN & 58.94 & 86.73 & 84.30 & 56.73 & 62.69 & 75.89 & 16.05 & 60.09 & 41.55 & 24.77 & 60.16 & 56.90 \\
DoTA & 61.41 & 87.51 & 85.01 & 61.37 & 60.78 & 73.81 & 18.63 & 63.21 & 41.61 & 23.67 & 62.28 & 57.79 \\
OGA & 59.30 & 85.80 & 84.27 & 58.19 & 63.38 & 75.06 & 15.57 & 61.67 & 41.84 & 34.43 & 61.12 & 58.13 \\
BCA & 58.97 & 86.57 & 82.53 & 57.95 & 58.26 & 68.50 & 17.04 & 59.65 & 38.89 & 10.28 & 59.34 & 53.90 \\
ZERO & 59.86 & 86.33 & 84.46 & 57.80 & 58.99 & 73.81 & 17.64 & 60.32 & 39.89 & 21.62 & 59.32 & 56.02 \\
OnZeta & 62.69 & 86.13 & 87.79 & 61.02 & 64.43 & 79.35 & 18.87 & 64.37 & 44.74 & 34.02 & 65.29 & 60.60 \\
ECALP & 61.25 & 89.74 & 84.79 & 59.22 & 65.29 & 78.22 & 17.82 & 63.15 & 45.04 & 36.79 & 65.42 & 60.55 \\
MTA & 60.21 & 87.22 & 84.85 & 58.72 & 60.98 & 75.72 & 18.15 & 60.75 & 40.48 & 22.56 & 60.85 & 57.03 \\
\bottomrule
\end{tabular}
\end{adjustbox}
\end{table*}

\noindent\textbf{Distribution:}
\begin{itemize}[leftmargin=*, parsep=0pt, itemsep=2pt, topsep=2pt]
    \item \textbf{DoTA} \cite{dota}: We initialize the class-wise distribution mean from CLIP text features and set the initial mean value to $1\times10^{-3}$. The covariance smoothing parameters are $\epsilon=1\times10^{-4}$ and $\sigma=2\times10^{-3}$. The online update momentum is $\eta=0.3$, and the prior blending weight is $\rho=0.02$.
    \item \textbf{OGA} \cite{oga}: We use a shot capacity of $8$ and confidence threshold $\tau=0.05$. The precision matrix is estimated with the Ridge-Moore-Penrose form, and mean normalization is disabled.
    \item \textbf{BCA} \cite{bca}: We use two confidence-controlled count priors. The first branch uses threshold $0.05$ with initialization count $20{,}000$, while the second uses threshold $0.65$ with initialization count $1$. The method operates on confidence-selected augmented predictions with selection ratio $0.1$.
    \item \textbf{OnZeta} \cite{onzeta}: We run OnZeta without prompt tuning using batch size $1$. The text and image temperatures are $\tau_t=0.01$ and $\tau_i=0.04$. The classifier update coefficient is $c_w=0.5$, the prior update coefficient is $c_r=20.0$, and the balancing parameters are $\alpha=1.0$ and $\beta=0.8$. We average predictions over $5$ random online orders and use seven hand-crafted prompt templates.

\end{itemize}

\begin{table}[h]
\centering
\small
\caption{\textbf{Effect of AugMix on TPT and TPS.} Accuracy (\%) comparison between base augmentations and AugMix.}
\label{tab:augmix_effect}
\begin{tabular}{ll cccccc}
\toprule
\textbf{Method} & \textbf{Aug. Strategy} & \textbf{I-A} & \textbf{Food101} & \textbf{I-R} & \textbf{I-V2} & \textbf{SUN397} & \textbf{StanfordCars} \\
\midrule
\multirow{2}{*}{TPT} 
& Base & \textbf{53.77} & \textbf{86.64} & \textbf{76.58} & \textbf{63.36} & 65.08 & \textbf{67.23} \\
& AugMix & 52.03 & 86.34 & 76.54 & 62.70 & \textbf{65.51} & 67.07 \\
\midrule
\multirow{2}{*}{TPS} 
& Base & \textbf{62.00} & \textbf{84.42} & \textbf{79.30} & \textbf{64.80} & 63.18 & \textbf{69.02} \\
& AugMix & 58.47 & 82.93 & 79.29 & 64.02 & \textbf{63.70} & 68.54 \\
\bottomrule
\end{tabular}
\end{table}

\subsubsection{Inference-Based Methods}

\noindent\textbf{View Aggregation:}
\begin{itemize}[leftmargin=*, parsep=0pt, itemsep=2pt, topsep=2pt]
    \item \textbf{ZERO} \cite{zero}: We use AugMix to generate 63 augmented views per image. The final prediction is obtained by retaining the top 30\% most confident views and aggregating their predictions.
    \item \textbf{MTA} \cite{mta}: We use 64-view AugMix batches without prompt optimization. MTA estimates a feature-space mode by alternating density-based mode updates and confidence-affinity reweighting for up to $5$ inner iterations. The local bandwidth is computed from the nearest 30\% pairwise feature distances. We set $\lambda_q=4.0$ for prediction-affinity weighting and $\lambda_y=0.2$ for the soft assignment temperature.

\end{itemize}

\noindent\textbf{Feature Modulation:}
\begin{itemize}[leftmargin=*, parsep=0pt, itemsep=2pt, topsep=2pt]
    \item \textbf{CALIP} \cite{calip}: We implement the parameter-free attention module without test-time optimization. The visual-guided and textual-blended logit weights are set to $\beta_2=1.0$ and $\beta_3=0.01$, respectively.
    \item \textbf{Panda} \cite{panda}: We adapt normalization parameters continuously with SGD at learning rate $1\times10^{-4}$. Negative data augmentation is generated by patch shuffling with patch size $32\times32$. The logit-level bias offset strength is set to $\beta=0.2$.
\end{itemize}

\begin{table}[h]
\centering
\small
\caption{\textbf{Learning Rate Sensitivity.} Accuracy (\%) of TPT and TPS on ImageNet-A under different learning rates (with 1 adaptation step).}
\label{tab:lr_sensitivity}
\begin{tabular}{l cccccc}
\toprule
\multirow{2}{*}{\textbf{Method}} & \multicolumn{6}{c}{\textbf{Learning Rate}} \\
\cmidrule(lr){2-7}
 & \textbf{0.0001} & \textbf{0.0005} & \textbf{0.001} & \textbf{0.005} & \textbf{0.01} & \textbf{0.05} \\
\midrule
TPT & 48.09 & 49.07 & 49.68 & 52.03 & \textbf{52.71} & 47.44 \\
TPS & 51.96 & 54.63 & 56.05 & 58.47 & 58.75 & \textbf{59.05} \\
\bottomrule
\end{tabular}
\end{table}

\section{Additional Experiments and Analyses}
\label{app:exp}

\subsection{Details on ImageNet-C}
\label{app:exp_imagenetc}

Table \ref{tab:imagenetc_vit} provides the detailed performance breakdown of various TTA methods on the 15 corruption types of ImageNet-C (severity level 5) using the default CLIP ViT-B/16 backbone. The results demonstrate that while parameter-based and state-based methods struggle to maintain stability across all corruption types, norm-layer tuning methods (e.g., DeYO, SAR) explicitly targeting feature statistics achieve the most consistent robustness against low-level visual perturbations.

\subsection{Experiments with ResNet-50 Backbone}
\label{app:exp_rn50}

To ensure our conclusions are not strictly bound to the ViT architecture, we replicated our evaluations across natural, fine-grained, and corruption shifts using the ResNet-50 (RN50) backbone. As shown in Table \ref{tab:rn50-natural} and Table \ref{tab:rn50-fine}, the overall performance of RN50 is inherently lower than ViT-B/16. However, the relative rankings among TTA paradigms remain largely consistent: state-based methods excel in semantic fine-grained datasets, lightweight prototype tuning performs well on natural shifts, and heavy parameter updating continues to show limited or negative returns compared to training-free approaches.

\begin{table}[h]
\centering
\small
\caption{\textbf{Scaling Adaptation Steps.} Accuracy (\%) and total inference time (seconds) on ImageNet-A as the number of per-sample gradient steps increases.}
\label{tab:step_scaling}
\begin{tabular}{ll cccccccc}
\toprule
\multirow{2}{*}{\textbf{Method}} & \multirow{2}{*}{\textbf{Metric}} & \multicolumn{8}{c}{\textbf{Adaptation Steps}} \\
\cmidrule(lr){3-10}
 & & \textbf{1} & \textbf{2} & \textbf{3} & \textbf{4} & \textbf{5} & \textbf{10} & \textbf{15} & \textbf{20} \\
\midrule
\multirow{2}{*}{TPT} 
 & Acc (\%) & 52.03 & 53.69 & 54.20 & 54.79 & 54.91 & 55.96 & \textbf{56.13} & 56.11 \\
 & Time (s) & 1,546 & 2,733 & 3,899 & 5,103 & 6,285 & 12,142 & 18,069 & 23,943 \\
\midrule
\multirow{2}{*}{TPS} 
 & Acc (\%) & 58.47 & 58.87 & 59.07 & 59.08 & 59.16 & 59.17 & 59.20 & \textbf{59.21} \\
 & Time (s) & 1,263 & 1,265 & 1,395 & 1,832 & 2,262 & 4,428 & 6,592 & 8,756 \\
\bottomrule
\end{tabular}
\end{table}

\subsection{Ablation on AugMix Augmentation}
\label{app:exp_augmix}

Many prompt-tuning baselines inherently incorporate the AugMix data augmentation strategy by default. However, as demonstrated in Table \ref{tab:augmix_effect}, applying AugMix provides marginal to no improvements across most distribution shifts. In several cases, such as on ImageNet-A and Food-101, it actually degrades the prediction accuracy compared to a standard random crop and flip baseline (Base). Despite this, we retained the AugMix setup in our primary evaluations to strictly align with the established experimental settings of prior works (e.g., C-TPT, A-TPT).

\begin{table}[h]
\centering
\small
\caption{\textbf{Impact of Confidence Filtering Strategies.} Top-1 Accuracy (\%) of TPT and TPS on ImageNet-A across different view selection ratios ($p$).}
\label{tab:confidence_filtering}
\begin{tabular}{ll ccccc}
\toprule
\multirow{2}{*}{\textbf{Criterion}} & \multirow{2}{*}{\textbf{Method}} & \multicolumn{5}{c}{\textbf{Selection Ratio ($p$)}} \\
\cmidrule(lr){3-7}
& & \textbf{1.0 (All)} & \textbf{0.75} & \textbf{0.5} & \textbf{0.25} & \textbf{0.1 (Top 10\%)} \\
\midrule
\multirow{2}{*}{\textbf{Entropy}}
& TPT & 49.07 & 49.65 & 50.24 & 51.55 & \textbf{52.03} \\
& TPS & 50.75 & 52.12 & 54.51 & 57.17 & \textbf{58.47} \\
\midrule
\multirow{2}{*}{\textbf{MSP}}
& TPT & 49.07 & 49.55 & 50.16 & 51.00 & \textbf{51.72} \\
& TPS & 50.75 & 51.72 & 53.23 & 55.61 & \textbf{56.89} \\
\bottomrule
\end{tabular}
\end{table}

\subsection{Detailed Analysis of Optimization Hyperparameters}
\label{app:exp_hyperparameters}

In Sec.~\ref{sec:tricks} of the main text, we illustrated the limitations of heavy parameter optimization in TTA4CLIP methods in Fig.~\ref{fig:lr_step}. Here, we provide the exact numerical data for the learning rate sensitivity and the scaling of adaptation steps on the ImageNet-A dataset using the default ViT-B/16 backbone.

Table \ref{tab:lr_sensitivity} details the accuracy of TPT and TPS across varying learning rates. While TPS shows a relatively monotonic increase before saturating, TPT exhibits high volatility, peaking at an accuracy of $52.71\%$ with a learning rate of $0.01$ but collapsing to $47.44\%$ at $0.05$. This confirms that without task-specific oracle tuning, parameter-based methods are highly sensitive to learning rate choices and prone to suboptimal performance.

Table \ref{tab:step_scaling} reports the performance and wall-clock inference time as the number of per-sample adaptation steps increases. The data clearly demonstrates the diminishing returns of prolonged optimization: pushing TPT from 1 to 20 steps yields a marginal $\sim 4.08\%$ accuracy gain but inflates the inference time by over 15$\times$ ($1,546$s to $23,943$s). Meanwhile, TPS effectively saturates after the very first step, with subsequent steps offering less than $0.8\%$ improvement despite a linear increase in computational cost.

\subsection{Impact of Confidence Filtering Strategies}
\label{app:exp_confidence}

As discussed in Sec. \ref{sec:tricks:evidence} and Fig.~\ref{fig:tpt_exp}(b), filtering out noisy augmented views based on model confidence is a critical driver for successful parameter adaptation. In Table \ref{tab:confidence_filtering}, we provide the detailed ablation results for TPT and TPS on ImageNet-A across different selection ratios ($p$). We compare two common confidence criteria: predictive entropy and Maximum Softmax Probability (MSP).

The results demonstrate a clear and monotonic performance gain as the selection becomes stricter: utilizing all augmented views ($p=1.0$) leads to suboptimal results due to the inclusion of low-quality or semantically distorted views. By enforcing a strict selection threshold (e.g., $p=0.1$, keeping only the top 10\% most confident views), both methods achieve their peak accuracy. Notably, Entropy-based selection consistently outperforms MSP-based selection across all ratios, suggesting that entropy is a more reliable proxy for identifying high-quality evidence in test-time augmented views.

\subsection{Results with Additional CLIP Backbones}
\label{app:additional-backbones}

Table~\ref{tab:additional-backbones} reports results on ImageNet-A with CLIP ViT-B/32 and ResNet-101. 
Together with the ViT-B/16 and ResNet-50 results in Tables~\ref{tab:main-results} and~\ref{tab:rn50-natural}, these results show that the relative rankings of the methods and paradigms remain largely stable. 
For example, lightweight prototype updates such as TPS and DPE still maintain their advantages.
Moreover, We surprisingly observe that, with nearly the same number of parameters, ViT-B/32 is clearly less capable than ViT-B/16 under distribution shifts. 
A possible reason is that at an input resolution of $224\times224$, ViT-B/16 produces a $14\times14$ patch grid, whereas ViT-B/32 produces a $7\times7$ grid.
Finer spatial representation may contribute to its stronger performance, as it enables the model to perceive fine-grained, local visual details and improves robustness to downstream distributions.

\begin{table}[h]
\centering
\small
\caption{\textbf{Additional CLIP backbone results on ImageNet-A, including ViT-B/32 and ResNet-101.} Note that CALIP does not support Non-ViT architecture.}
\label{tab:additional-backbones}
\begin{tabular}{lcc}
\toprule
\textbf{Method} & \textbf{ViT-B/32} & \textbf{ResNet-101} \\
\midrule
CLIP & 29.51 & 28.33 \\
\midrule
TPT & 33.13 & 31.40 \\
DiffTPT & 32.04 & 31.89 \\
HisTPT & 29.05 & 28.40 \\
C-TPT & 30.52 & 28.52 \\
A-TPT & 31.85 & 30.19 \\
R-TPT & 18.72 & 15.99 \\
TPS & 39.81 & 35.80 \\
DPE & 38.49 & 34.71 \\
BATCLIP & 31.63 & 12.21 \\
RLCF & 30.55 & 29.61 \\
\midrule
TDA & 30.69 & 29.47 \\
BoostAdapter & 30.71 & 29.19 \\
DMN & 29.81 & 28.60 \\
ECALP & 31.12 & 29.11 \\
DoTA & 39.73 & 36.68 \\
OGA & 30.03 & 28.85 \\
BCA & 38.41 & 35.05 \\
OnZeta & 31.17 & 29.77 \\
\midrule
ZERO & 37.37 & 34.43 \\
MTA & 36.83 & 33.67 \\
Panda & 30.51 & 13.77 \\
CALIP & 31.37 & - \\
\bottomrule
\end{tabular}
\end{table}

\subsection{Effect of Test-Stream Ordering}
\label{app:stream-order}

State-based methods accumulate cross-sample evidence, which may depend on the order in which samples arrive. 
To assess this effect, we rank each test image offline by the entropy of its zero-shot CLIP prediction, using low entropy as a proxy for an easy sample. 
We compare a random order with easy-to-hard and hard-to-easy orders on ImageNet-A and Food101, using ViT-B/16 backbone. 
Each method starts from its original initialization for each stream.

\begin{table}[h]
\centering
\small
\caption{\textbf{Effect of test-stream order.} Top-1 accuracy (\%) of four representative state-based methods, including cache-based TDA and ECALP, and distribution-based DoTA and OnZeta.}
\label{tab:stream-order}
\begin{tabular}{llcccc}
\toprule
\textbf{Dataset} & \textbf{Order} & \textbf{TDA} & \textbf{ECALP} & \textbf{DoTA} & \textbf{OnZeta} \\
\midrule
\multirow{3}{*}{ImageNet-A}
& Random       & 48.83 & 49.12 & 58.43 & 48.55 \\
& Easy-to-hard & \textbf{49.51} & 49.36 & \textbf{58.78} & 48.57 \\
& Hard-to-easy & 48.11 & \textbf{49.85} & 58.26 & \textbf{48.79} \\
\midrule
\multirow{3}{*}{Food101}
& Random       & \textbf{85.64} & 87.14 & 85.13 & 87.59 \\
& Easy-to-hard & 85.55 & \textbf{87.34} & \textbf{85.21} & \textbf{87.60} \\
& Hard-to-easy & 85.22 & 85.65 & 84.92 & 87.57 \\
\bottomrule
\end{tabular}
\end{table}

Table~\ref{tab:stream-order} demonstrates the results of four representative state-based methods. 
Generally, for state-based methods, going from easy to hard streams indeed works better than going from hard to easy streams. 
This may be because early high-confidence samples help build more reliable caches or class statistics.
Overall, the default random ordering remains relatively robust for TTA4CLIP methods. 

\subsection{Evidence Quality and Forgetting in Online Adaptation}
\label{app:forgetting-evidence}

To examine how evidence quality affects continual parameter updates, we study TPT~\cite{tpt} on ImageNet-A with ViT-B/16 and disable episodic prompt resets ($\tau=\infty$). 
For each image, we generate 64 augmented views and use their pre-update predictions to construct an oracle-selected evidence set. 
The oracle uses ground-truth labels only to determine whether each view is predicted correctly, and the model is still updated using TPT's unsupervised entropy objective. 

We then replace a fraction $r\in\{0,0.1,0.2,0.5\}$ of the correctly predicted views with incorrectly predicted views from the same image. 
The requested number of replacements is $\lfloor r n_c\rfloor$, where $n_c$ is the number of correctly predicted views, capped by the number of available incorrect views. 
We also report standard online TPT and HisTPT~\cite{histpt}, which uses historical evidence, as references.

\begin{table*}[t]
\centering
\small
\caption{\textbf{Evidence quality and continual adaptation.} ImageNet-A top-1 accuracy (\%) in consecutive 500-sample logging windows. TPT$^\star$ uses oracle-selected views and target replacement fraction $r$. TPT and HisTPT are both performed online without episodic resets, as in Fig.~\ref{fig:tpt_exp}(d).}
\label{tab:forgetting-evidence}
\begin{adjustbox}{width=\textwidth}
\begin{tabular}{llccccccccccccccc}
\toprule
\textbf{Method} & $\boldsymbol{r}$ & \multicolumn{15}{c}{\textbf{Window Order during Test-Time}} \\
\cmidrule(lr){3-17}
& & 1 & 2 & 3 & 4 & 5 & 6 & 7 & 8 & 9 & 10 & 11 & 12 & 13 & 14 & 15 \\
\midrule
TPT & -- & 50.60 & 41.20 & 17.40 & 10.80 & 8.80 & 6.80 & 6.57 & 9.27 & 5.79 & 6.57 & 2.85 & 5.55 & 5.64 & 3.18 & 3.33 \\
TPT$^\star$ & 0 & 64.10 & 66.70 & 61.40 & 61.40 & 59.80 & 64.40 & 61.80 & 58.80 & 59.40 & 60.00 & 65.40 & 63.80 & 61.80 & 62.80 & 65.80 \\
& 0.1 & 63.80 & 64.00 & 60.40 & 59.80 & 59.20 & 61.00 & 60.00 & 57.00 & 58.00 & 59.40 & 64.20 & 63.60 & 61.20 & 60.40 & 64.80 \\
& 0.2 & 64.20 & 62.60 & 59.40 & 58.40 & 56.40 & 59.60 & 59.60 & 57.00 & 56.60 & 58.40 & 61.80 & 62.80 & 60.40 & 59.60 & 63.60 \\
& 0.5 & 29.40 & 1.80 & 2.00 & 1.60 & 1.20 & 1.60 & 2.20 & 1.60 & 0.60 & 4.00 & 3.20 & 1.80 & 2.60 & 1.00 & 2.60 \\
HisTPT & -- & 49.80 & 49.40 & 46.80 & 46.00 & 45.80 & 49.00 & 48.40 & 44.40 & 47.20 & 48.20 & 46.80 & 47.80 & 46.40 & 47.60 & 51.80 \\
\bottomrule
\end{tabular}
\end{adjustbox}
\end{table*}

Table~\ref{tab:forgetting-evidence} shows that: 
(i) Fully correct oracle-selected evidence does not cause catastrophic forgetting. Instead, it supports stable continual adaptation and yields substantial performance gains.
(ii) The model can still adapt continually with a limited proportion of incorrect evidence $r$, but it rapidly collapses once the proportion reaches a certain level (e.g., $r=0.5$).
(iii) An explicit retention mechanism for cross-sample evidence can support sustained evidence utilization while avoiding severe forgetting.

%%%%%%%%%%%%%%%%%%%%%%%%%%%%%%%%%%%%%%%%%%%%%%%%%%%%%%%%%%%%

% \newpage
% \input{checklist.tex}

\end{document}